\documentclass[10pt,twocolumn,letterpaper]{article}
\usepackage[pagenumbers]{cvpr}
\usepackage[utf8]{inputenc}
\usepackage[T1]{fontenc}
\usepackage{graphicx,booktabs,multirow,amsmath,amssymb,nicefrac}
\usepackage{microtype}
\usepackage[most]{tcolorbox}
\usepackage{capt-of,placeins,float,url}
\usepackage{stfloats} 

\definecolor{cvprblue}{rgb}{0.21,0.49,0.74}
\usepackage[pagebackref,breaklinks,colorlinks,allcolors=cvprblue]{hyperref}
\def\confName{CVPR}
\def\confYear{2027}
\title{Ranking-PE: Ranking-Aware Prompt Evolution\\for Clinical Vision--Language Models}
\author{%
\setlength{\tabcolsep}{6pt}
\begin{tabular}{@{}ccccc@{}}
Tian Xia$^{1}$ &
Minghao Liu$^{2}$ &
Yiqing Liang$^{3}$ &
Laixi Shi$^{4}$ &
Jiayun Wang$^{5}$
\\[2pt]
\multicolumn{5}{c}{\normalfont\footnotesize
  $^1$Harvard University \;
  $^2$University of California, Santa Cruz \;
  $^3$Brown University}
\\[2pt]
\multicolumn{5}{c}{\normalfont\footnotesize
  $^4$Johns Hopkins University \;
  $^5$Georgia Institute of Technology}
\end{tabular}
}
\hypersetup{pdftitle={Ranking-PE: Ranking-Aware Prompt Evolution for Clinical Vision--Language Models},pdfauthor={Tian Xia, Minghao Liu, Yiqing Liang, Laixi Shi, Jiayun Wang}}
\makeatletter
\patchcmd{\@maketitle}{\vspace*{24pt}}{\vspace*{12pt}}{}{\errmessage{Title spacing patch failed}}
\patchcmd{\@maketitle}{\vskip .375in}{\vskip 0in}{}{\errmessage{Top title spacing patch failed}}
\makeatother
\begin{document}
\maketitle
\begin{abstract}
Multimodal large language models (MLLMs) are rapidly advancing clinical diagnosis, yet their adaptation pipelines remain anchored to accuracy-based objectives. Clinical data are heavily class-imbalanced: a constant-majority predictor can score above $90\%$ accuracy while being clinically useless. We therefore evaluate diagnostic ranking using AUROC, a threshold-free measure of how well positives rank above negatives that is invariant to class balance.
We focus on reflective prompt optimization in MLLMs. In accuracy-directed Pareto search, each row of the binary scores matrix represents an evaluation instance and each column a candidate prompt. Cells record per-instance correctness, so the column average is accuracy and drives candidate selection.

We introduce \emph{pair-level Pareto prompt evolution} (Ranking-PE), which replaces instance-correctness rows in the scores matrix with pairwise-ordering rows. Each cell evaluates whether a candidate prompt ranks a positive example above its paired negative, making the column average equal to empirical AUROC while preserving pair-level differences between candidates. We reformulate all three layers that guide prompt evolution with the ranking objective: candidate comparison using this matrix, feedback for prompt revision, and final candidate selection---with no extra model calls and no surrogate loss.
Across three diseases on MIMIC, accuracy-based prompt evolution can degrade diagnostic ranking; Ranking-PE reverses this, outperforming the accuracy-based recipe by $+5.8$ AUROC pp on fine-tuned Qwen3-VL-8B and $+16.2$ pp on MedGemma-4B. Ablations examine each design component and highlight the importance of medical domain adaptation through vision-encoder tuning or medical pretraining. Our recipe extends reflective prompt evolution from text-only data to multimodal clinical decision-making.
\textnormal{\textbf{Project page:} \url{https://ranking-pe.github.io/}}
\vspace{-20pt}
\end{abstract}

\section{Introduction}
\label{sec:intro}

Multimodal large language models (MLLMs) are being developed for
multimodal clinical diagnosis, jointly reading a medical image (e.g., chest
X-ray) and the accompanying structured EHR context (demographics, imaging
view, laboratory test results) to produce a diagnostic
answer~\citep{tu2024medpalmm, li2023llavamed, google2025medgemma}.
The promise is that a single generalist model can approach specialist-level
decisions without disease-specific architectures. Shared pretrained knowledge offers a route to adapting across diseases and tasks with limited disease-specific supervision. Their language interface also allows diagnostic rationales to be presented for clinician review. Recent medically pretrained
backbones, including LLaVA-Med~\citep{li2023llavamed},
Med-PaLM~M~\citep{tu2024medpalmm}, and MedGemma~\citep{google2025medgemma},
represent progress toward this goal.

\begin{figure*}[t]
    \centering
    \includegraphics[width=0.8\textwidth]{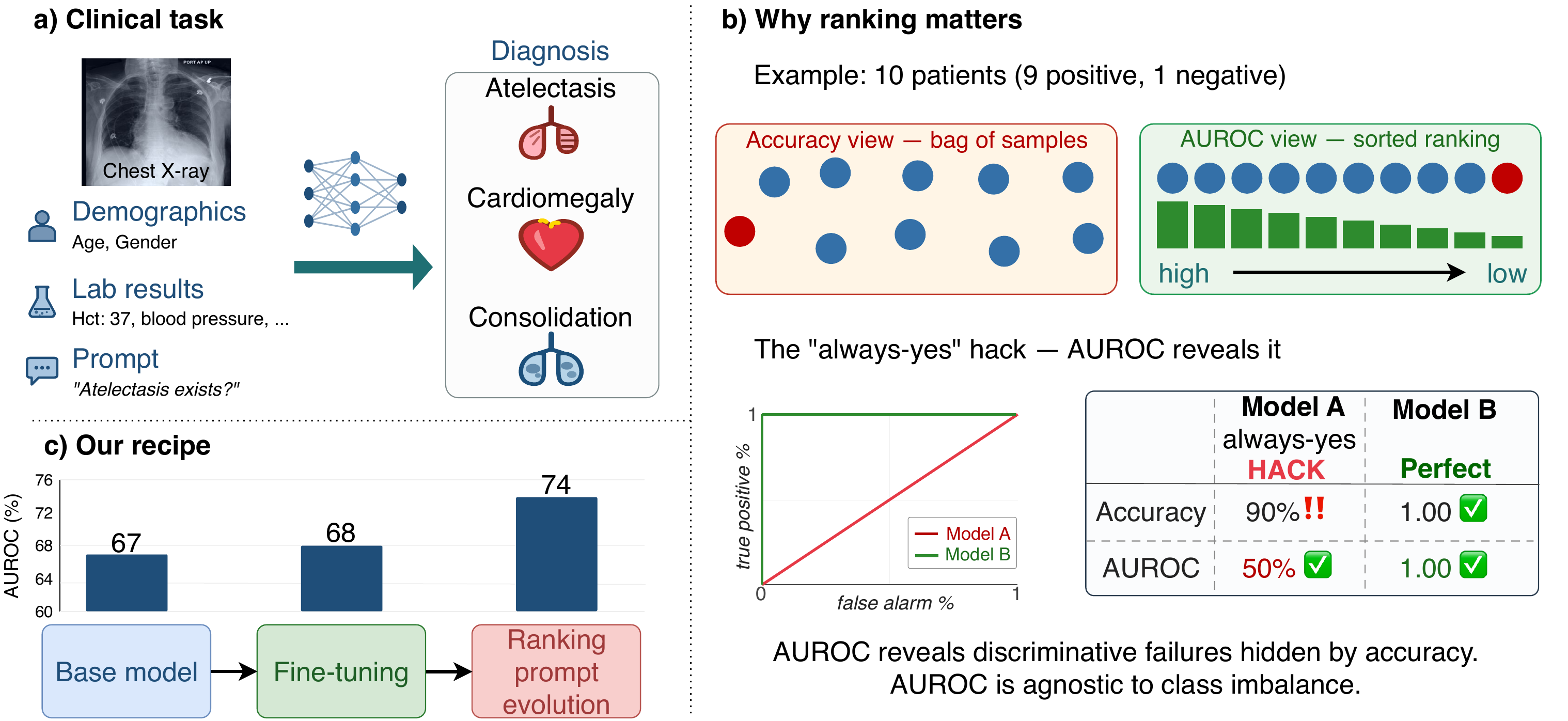}
    
    \caption{\textbf{Ranking-aware prompt evolution for clinical diagnosis.}
    \emph{(a)} An MLLM combines a chest X-ray with clinical context and a task prompt to produce a diagnostic answer and score.
    \emph{(b)} Under class imbalance, high accuracy can conceal poor discrimination between positive and negative cases.
    \emph{(c)} Our recipe combines supervised fine-tuning (SFT) with ranking-aware prompt evolution. Bars show rounded mean test AUROC on Qwen3-VL-8B across the three disease tasks, before adaptation, after SFT, and after SFT followed by Ranking-PE.
    }
    \label{fig:overview}
\end{figure*}

Accuracy alone is not enough for clinical diagnosis model training and evaluation as it does not consider the imbalanced nature of clinical data.
Screening, triage, and follow-up operate at chosen points on the receiver
operating characteristic (ROC) curve, which plots the true positive rate against
the false positive rate as the decision threshold varies. Diagnostic ranking is
therefore a central concern: the area under this curve (AUROC) measures
discrimination across thresholds, whereas accuracy measures label agreement at a
single operating point.
Under severe class imbalance, a constant-majority predictor can achieve over $90\%$ accuracy without distinguishing positive from negative cases, missing every positive case when negatives dominate or flooding clinicians with false alarms when positives dominate (Figure~\ref{fig:overview}b).
Errors are also asymmetric: a missed consolidation is clinically far more
costly than a false alarm, a distinction accuracy is structurally incapable of
representing.
Under these conditions, optimizing and reporting accuracy actively
\emph{misdirects} MLLM development in medicine; we therefore align prompt evolution with diagnostic ranking and evaluate its
performance using AUROC, which is threshold-free and agnostic to class imbalance. AUROC itself weighs both error types symmetrically;
we surface the asymmetric cost separately, through the reflector's feedback
(\S\ref{sec:feedback_v2}).

This paper focuses on prompt optimization \citep{agrawal2025gepa, khattab2024dspy, pryzant2023apo} (cheap,
gradient-free, and composable with finetuning techniques) layered on top of a finetuned
medical backbone.
The central challenge is to make diagnostic ranking guide the search that produces a prompt.
Existing prompt-evolution systems (GEPA \citep{agrawal2025gepa}, DSPy \citep{khattab2024dspy}, APO \citep{pryzant2023apo}) are built around an
\emph{accuracy-based scores matrix}, inheriting exactly the metric
misalignment described above. Under heavy class imbalance, accuracy-based
prompt evolution can degrade the underlying ranking signal, a failure mode
our recipe reverses.
Pareto prompt evolution does not intrinsically optimize accuracy; its
comparisons depend on the scores assigned to each row of its matrix.
Existing systems silently choose per-\emph{instance} correctness events
$\mathbf{1}[\text{correct}]$ (cell in the score matrix is 1 when correct), making the column average accuracy.
For clinical ranking, we instead score positive-negative pairs according to their relative order. We obtain the per-instance scores $s(\cdot)$ from the model's Yes/No token log-probabilities at no additional cost.
By the Wilcoxon--Mann--Whitney identity, the average of these pair scores is
empirical AUROC.
Candidates can correctly order different pairs, and the pair-level matrix preserves
these differences. Swapping instance rows for $(x^+, x^-)$ pairs makes
candidate comparison and selection depend on
pairwise ranking, while final selection maximizes validation AUROC.
We call this {\it pair-level Pareto prompt evolution} (Ranking-PE).

We consider MLLM-based clinical diagnosis as a \emph{ranking task} and validate
the recipe systematically end-to-end across two open-source MLLM families
(Qwen3-VL-8B and MedGemma-4B) over three disease tasks on MIMIC,
reporting AUROC and balanced accuracy in the main paper and AUPRC in the
appendix; a same-scale non-medical checkpoint
(Gemma3-4B) and frontier closed-source GPT reflectors probe the roles of medical
adaptation and reflector choice, respectively.
Ranking-PE outperforms accuracy-based prompt evolution by $+5.8$ AUROC pp on
Qwen3-VL-8B with vision-encoder-tuned SFT and $+16.2$ AUROC pp on MedGemma-4B,
while exceeding the balanced accuracy of both Accuracy-PE and the corresponding
base model.
On Qwen3-VL-8B + SFT, the pattern is most pronounced on Atelectasis,
where Accuracy-PE stays close to the base level;
Ranking-PE improves AUROC by 7.1 pp and balanced accuracy by 25.5 pp over Accuracy-PE under the same pipeline.

Our contribution: 
    1) \textbf{Pair-level Pareto prompt evolution.}
    We identify the row schema of the Pareto scores matrix as the
    \emph{hidden optimization unit} of reflective prompt search, and
    reformulate all three layers the search reads from (the scores matrix,
    reflector feedback, and final selection) around positive-negative
    ordering events $(x^+, x^-)$, so that ranking governs both which
    candidates evolve and how their prompts are revised (\S\ref{sec:method}).
    2) \textbf{A systematic clinical recipe and benchmark.}
    We provide an end-to-end study of MLLM-based clinical diagnosis on
    MIMIC-IV (three diseases, two open-source MLLM families, four
    optimization paradigms; closed-source models enter only as probes,
    namely zero-shot baselines and reflectors), show that visual domain adaptation provides an important foundation
    for prompt evolution (consistent with
    prior visual-alignment findings~\citep{li2023llavamed,he2024pefomed}), and extend
    reflective prompt evolution from text-only NLP benchmarks to
    multimodal clinical decision-making (\S\ref{sec:experiments}).
    

\section{Related Work}
\label{sec:related}

\noindent \textbf{Multimodal LLMs for medicine and adaptation.}
Multimodal LLMs specialized to the medical domain, including LLaVA-Med~\citep{li2023llavamed}, Med-PaLM~M~\citep{tu2024medpalmm}, BiomedGPT~\citep{zhang2024biomedgpt}, and MedGemma~\citep{google2025medgemma}, are typically built from general-purpose backbones by continued pretraining and instruction tuning for medical VQA and report generation. Their adaptation pipelines often use parameter-efficient methods such as LoRA~\citep{hu2022lora} and QLoRA~\citep{dettmers2023qlora}. However, downstream accuracy alone is ill-suited to the heavy class imbalance pervasive in clinical data: it leaves rank quality between positives and negatives unassessed. We instead evolve prompts to improve ranking and use AUROC for evaluation; our experiments show that SFT alone remains limited on image-dependent diseases, while ranking-aware prompt evolution substantially improves diagnostic ranking.

\noindent \textbf{Ranking-aware learning and AUROC optimization.}
Direct optimization of ranking metrics has a long history in learning to rank: RankNet~\citep{burges2005ranknet}, LambdaRank~\citep{burges2006lambdarank}, and AUC-maximization on medical imaging~\citep{yuan2021aucmax} train new parameters against differentiable surrogates of pairwise objectives. These methods must replace the indicator $\mathbf{1}[s(x^+)>s(x^-)]$ with logistic, hinge, or sigmoid-based surrogates because the indicator provides no useful gradient almost everywhere. This requirement arises from gradient-based optimization, rather than from the ranking metric itself. In the clinical setting, threshold-free reporting standards (TRIPOD~\citep{collins2015tripod}, CheXpert sensitivity at fixed specificity~\citep{irvin2019chexpert}) motivate ranking-first evaluation regardless of training. Gradient-free prompt search can therefore use the pairwise ordering indicator directly, without introducing a surrogate loss.

\noindent \textbf{Prompt optimization for LLMs.}
A growing line of work optimizes text prompts without updating weights: APO~\citep{pryzant2023apo} performs gradient-free edits via textual ``gradients''; DSPy~\citep{khattab2024dspy} compiles declarative pipelines with prompt-level teleprompters; PromptBreeder~\citep{fernando2023promptbreeder} uses self-referential evolutionary search; GEPA~\citep{agrawal2025gepa} maintains a Pareto frontier of candidates refined by reflective LLM feedback; in medicine, Medprompt~\citep{nori2023medprompt} composes prompting strategies for accuracy on medical QA. These methods read binary per-instance signals into a scores matrix that consumes indicators directly and thus requires no smoothed surrogate. By aligning the row event with pairwise ordering rather than instance correctness (\S\ref{sec:method}), the optimized quantity becomes the unrelaxed pairwise ranking objective itself. This requires no continuous surrogate, auxiliary loss, or extra rollouts.

\begin{figure*}[t]
    \centering
    \includegraphics[width=0.8\linewidth]{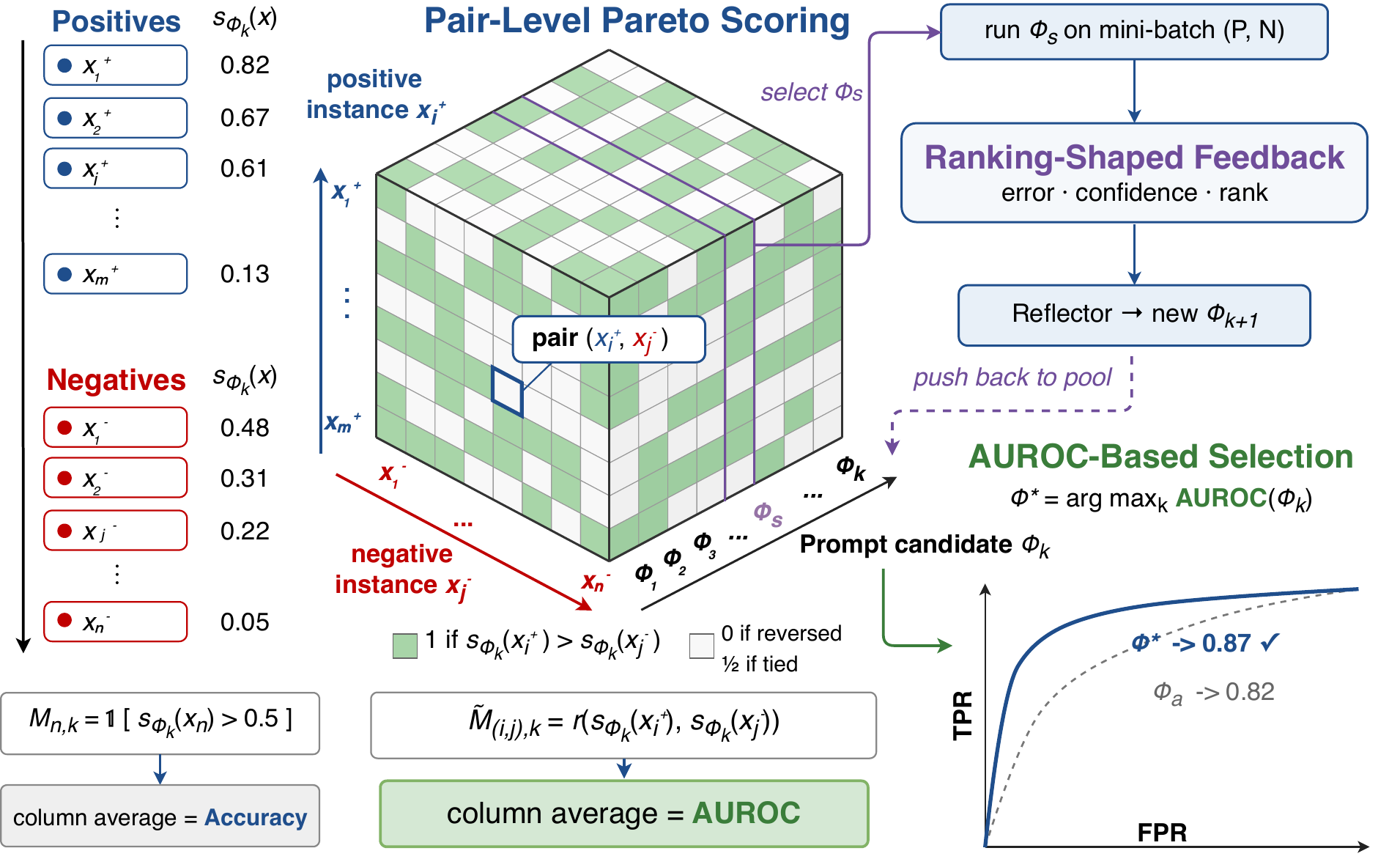}

    \caption{\textbf{Ranking-PE guides candidate comparison, prompt revision, and final selection.}
    \emph{(Left)} Positives and negatives denote samples with positive and negative ground-truth labels, respectively. The pair-level matrix retains which positive--negative pairs each candidate orders correctly for Pareto comparison and parent selection beyond a single AUROC value. Cells score correct, tied, and reversed orderings as $1$, $0.5$, and $0$; the mean over pairs equals empirical AUROC.
    \emph{(Right top)} Ranking-shaped feedback guides the reflector with clinical error type, confidence magnitude, and rank context. Revised prompts return to the candidate pool.
    \emph{(Right bottom)} Final selection returns the candidate with the highest validation AUROC.
    }
    \label{fig:method}
\end{figure*}
\section{Task: Clinical Diagnosis as Ranking}
\label{sec:task_v2}

We study binary clinical diagnosis from paired inputs
$x = (\mathtt{image}, \mathtt{ehr})$, where $\mathtt{ehr}$ is a structured
electronic-health-record context in text format (e.g.\ patient demographics, imaging view, and
laboratory values), together with a target disease $d$.
Let $\theta$ denote the (optional) fine-tuned adapter weights of a
multimodal language model and $\Phi$ a prompt that casts the task as a
templated question (``\emph{Does this patient have disease $d$?}'').
The model's output is a distribution over the two label tokens:
$p_{\theta, \Phi}(\hat{y} \mid x, d), \quad \hat{y} \in \{\mathtt{Yes}, \mathtt{No}\}$.
For brevity we suppress $\theta$ when held fixed and write $p_\Phi$
for the output distribution and $\hat{y}$ for the emitted token.

\noindent \textbf{Continuous score from a Yes/No head.}
Any ranking metric requires a continuous score, yet the model emits a discrete label. We recover a score from the logprobs a standard decoding pass already produces at no additional rollout cost: at the answer-token position we read the top-$k$ logprobs and define the log-odds score~\citep{holtzman2021surfaceform,kadavath2022pknow}
\begin{equation}
    s_\Phi(x, d) \;=\; \log p_\Phi(\mathtt{Yes} \mid x, d)
                    \;-\; \log p_\Phi(\mathtt{No} \mid x, d).
    \label{eq:score_v2}
\end{equation}
For brevity we write $s_\Phi(x)$ when $d$ is held fixed. During prompt evolution, the model weights remain fixed, and we search for prompts that improve the ordering induced by $s_\Phi$. $s_\Phi(x)$ provides a shared score for the scores matrix, per-example feedback, and final selection. Tokenization, top-$k$, label-fallback handling, and serving details are deferred to Appendix~\ref{sec:appendix_impl}.

\noindent \textbf{Performance metric: AUROC.}
Screening, triage, and confirmation impose different prevalences and cost ratios~\citep{pepe2003statistical,wynants2020covidreview}, so the clinical operating point is chosen at deployment. The right evaluation summary therefore does \emph{not} commit to a single threshold up front. Let $P$ and $N$ denote the positive and negative examples for target disease $d$, and let $X^+ \in P$, $X^- \in N$ denote a randomly drawn positive and negative example. AUROC is the probability that $s_\Phi$ ranks $X^+$ above $X^-$:

\begin{equation}
    \begin{aligned}
    \mathrm{AUROC}(\Phi) &= \Pr\!\left(s_\Phi(X^+, d) > s_\Phi(X^-, d)\right) \\
    &\quad + \tfrac{1}{2}\Pr\!\left(s_\Phi(X^+, d) = s_\Phi(X^-, d)\right),
    \end{aligned}
    \label{eq:auroc_integral}
\end{equation}
a threshold-free measure of whether the model ranks diseased patients above non-diseased ones. AUROC is the metric that ROC-based clinical reporting standards (TRIPOD~\citep{collins2015tripod}, CheXpert~\citep{irvin2019chexpert}) recommend. We use AUROC to evaluate diagnostic ranking and select the final prompt; \S\ref{sec:method} describes how pairwise ordering guides prompt evolution. AUPRC provides complementary information about precision and recall~\citep{saito2015prplot,davis2006prroc}, and we report it in Appendix Table~\ref{tab:main_full}. We additionally report balanced accuracy at the default decoder threshold to assess whether improved ranking is accompanied by better binary diagnostic decisions.

\section{Method} 
\label{sec:method}
\label{sec:method_v2}

Our method has two stages. First, a domain-adaptation supervised fine-tuning (SFT) gives the model a medical-image-aware visual backbone that produces a discriminative score $s_\Phi(x)$ (\S\ref{sec:warmup}). Second, a training-free \emph{ranking-aware prompt evolution} step refines the instruction prompt to improve diagnostic ranking by reformulating candidate comparison, feedback for prompt revision, and final candidate selection around the ranking objective (\S\ref{sec:stage2}).

\subsection{SFT with Vision-Encoder Tuning}
\label{sec:warmup}

A visual backbone adapted to the medical domain provides an important foundation for discriminative scores $s_\Phi(x)$. Prompt optimization alone may not fully compensate for limited domain adaptation, so Stage~1 uses SFT to strengthen this foundation. Given $\mathcal{D}_\mathrm{SFT} = \{(x_n, d_n, y_n)\}_{n=1}^{L}$ with $x_n = (\mathrm{image}, \mathrm{ehr})$, target disease $d_n$, and ground-truth label token $y_n \in \{\mathtt{Yes}, \mathtt{No}\}$, we minimize the supervised next-token negative log-likelihood
$\mathcal{L}_\mathrm{SFT}(\theta) = -\tfrac{1}{L}\sum_{n=1}^{L}\log p_\theta\!\bigl(y_n \mid x_n, d_n\bigr)$,
with $\theta$ comprising LoRA adapters on the language backbone and the full vision-encoder weights. Stage~2 holds the adapted model fixed and evolves only the prompt. Stage~1 is skipped when the checkpoint is already medically pretrained (e.g., MedGemma); the recipe then runs Stage~2 directly. Full hyperparameters are in Appendix~\ref{sec:appendix_impl}.

\subsection{Ranking-Aware Prompt Evolution}
\label{sec:stage2}

Ranking-PE uses pairwise ranking information to guide prompt evolution.
Final selection determines which prompt is returned, but the candidate pool is itself shaped by Pareto bookkeeping and reflector feedback, which under Accuracy-PE remain accuracy-steered. We therefore align both components with the ranking objective. After reviewing reflective Pareto-based prompt evolution (\S\ref{sec:prelim_gepa}), we introduce the pair-level scores matrix for comparing candidates through pairwise ordering (\S\ref{sec:pair_pareto_v2}) and ranking-shaped per-example feedback to guide prompt revision (\S\ref{sec:feedback_v2}); \S\ref{sec:pipeline_v2} composes these components with final selection into the complete search procedure.

\noindent \textbf{Background: reflective Pareto-based prompt evolution.}
\label{sec:prelim_gepa}
Reflective Pareto-based prompt evolution~\citep{agrawal2025gepa} is a training-free optimizer that searches for prompts improving task performance without updating model weights. Candidate prompts $\Phi_k$ are evaluated on a shared held-out set $\mathcal{V}$ of instances, producing per-instance predictions $\hat{y}_{n,k}$. In Accuracy-PE, correctness enters a binary \emph{scores matrix} $M \in \{0,1\}^{|\mathcal{V}| \times K}$ with $M_{n,k} = \mathbf{1}[\Phi_k \text{ correct on } n]$, whose column average is the accuracy of $\Phi_k$. The search filters out Pareto-dominated candidates (those whose correct-instance set is a strict subset of another's) and samples a parent candidate. A frozen \emph{reflector} reads textual diagnostics $\mu_f(x_n, \hat{y}_{n,k}, y_n)$ from a sampled minibatch to propose a revised prompt. After the budgeted rounds, the optimizer returns the validation $\arg\max_k$ over column averages of $M$.

The scores matrix records a binary event per row; its rows define Pareto dominance, and its column average is the quantity final selection maximizes. We identify this row schema as the search's \emph{hidden optimization unit} and specialize it to a pairwise-ordering event $(x^+, x^-)$, redirecting the column average to empirical AUROC. The matrix specialization redefines the search objective around pairwise ranking and preserves which pairs each candidate correctly orders. Final selection and reflector feedback align the remaining search components with this ranking objective.


\subsubsection{Pair-Level Pareto Scoring}
\label{sec:pair_pareto_v2}

To guide candidate comparison by pairwise ranking, we change the row schema of the scores matrix from instance correctness to the natural pairwise event for binary ranking: \emph{a positive scored above a negative}.

\noindent \textbf{The pair-indexed matrix.}
Let $P$ and $N$ be the positive and negative instances in the held-out split, with $x_i^+$ the $i$-th positive and $x_j^-$ the $j$-th negative. Define the pair score as $r(a,b)=\mathbf{1}[a>b]+\tfrac{1}{2}\mathbf{1}[a=b]$. The Wilcoxon--Mann--Whitney identity~\citep{hanley1982auroc} expresses empirical AUROC (Eq.~\ref{eq:auroc_integral}) as the average of these scores:
\begin{equation}
    \mathrm{AUROC}(\Phi) \;=\;
    \frac{1}{|P|\,|N|}
    \sum_{(i,j) \in P \times N}
    r\bigl(s_\Phi(x_i^+),s_\Phi(x_j^-)\bigr).
    \label{eq:pair_auroc_v2}
\end{equation}
We define the pair-indexed matrix $\tilde{M} \in \{0,\tfrac{1}{2},1\}^{(|P|\cdot|N|)\times K}$ by $\tilde{M}_{(i,j),k}=r(s_{\Phi_k}(x_i^+),s_{\Phi_k}(x_j^-))$. Over all positive-negative pairs, its column mean equals validation AUROC by Eq.~\ref{eq:pair_auroc_v2}, so final selection in Eq.~\ref{eq:final_selection} maximizes validation AUROC. A candidate Pareto-dominates another if its score is at least as high on every pair and strictly higher on at least one pair. This replaces dominance based on instance correctness with dominance based on pairwise ranking scores. Candidates that excel on different pairs can remain non-dominated, so the frontier can grow large. The selector samples frontier candidates in proportion to the number of pairs on which they achieve the highest score~\citep{agrawal2025gepa}, while final selection maximizes the column mean. This reformulation makes pairwise ordering a direct driver of prompt evolution through both candidate comparison and parent selection. Reflector feedback flows through a separate textual channel and needs its own alignment (\S\ref{sec:feedback_v2}).

\noindent \textbf{An unrelaxed pairwise ranking objective.}
Each cell records the exact pair score: $1$ for correct order, $\tfrac{1}{2}$ for equal scores, and $0$ for reversed order. This tie convention matches Eq.~\ref{eq:auroc_integral}. The optimizer uses these values directly without gradients, so no smoothing is needed. The optimized quantity is the empirical pairwise ranking objective itself, rather than a differentiable surrogate of the type used in classical learning-to-rank methods (RankNet~\citep{burges2005ranknet}, LambdaRank~\citep{burges2006lambdarank}, AUC-maximization~\citep{yuan2021aucmax}); see \S\ref{sec:related}.

\noindent \textbf{Cost.}
Constructing $\tilde{M}$ adds \emph{zero rollouts} over standard accuracy-based reflective search: pair scores are computed by table lookup from the $O(|P|+|N|)$ per-instance scores $\{s_{\Phi_k}(x)\}$ the standard validation pass produces. The matrix holds $|P|\cdot|N|$ score entries per candidate, assembled in $O(|P|\cdot|N|)$ CPU lookups. For larger splits we uniformly subsample rows to a cap $\texttt{max\_pairs}$; for any fixed candidate, the sampled column mean is an \emph{unbiased Monte Carlo estimator} of its empirical AUROC. Robustness to the cap is reported in Appendix~\ref{sec:appendix_impl}.
\subsubsection{Ranking-Shaped Per-Example Feedback}
\label{sec:feedback_v2}

The pair-level matrix guides candidate comparison and parent selection through ranking events, but the reflector's textual diagnostic $\mu_f$ still surfaces correctness-shaped signal without explicit information about score ordering. We therefore recast reflector feedback as a ranking-aware critique that directly guides prompt revision, extending ranking-aware search from candidate selection to candidate generation. Vanilla $\mu_f(x_n, \hat{y}_{n,k}, y_n)$ reduces to $\mathbf{1}[\hat{y}_{n,k} = y_n]$, discarding everything $s_{\Phi_k}(x_n)$ already tells us about confidence and its position relative to a reference validation score distribution. We extend $\mu_f$ with three signals, each rendered as a short sentence in the feedback read by the reflector. \emph{(i)} A \emph{clinical error type} $e \in \{\mathrm{TP},\mathrm{TN},\mathrm{FN},\mathrm{FP}\}$, with FN annotated as ``clinically dangerous'' to surface the asymmetric error costs of clinical screening. \emph{(ii)} A \emph{confidence magnitude}: the signed $s_{\Phi_k}(x_n)$ together with a plain-language bucket (low/moderate/high/extremely high). \emph{(iii)} A \emph{cross-example rank context}. This signal is computed from cached per-instance validation scores and can be used independently of the pair-level Pareto matrix. Let $s_{\mathrm{ref}}$ denote validation scores from the most recently fully evaluated candidate. For $x_n \in P$, we compare its current score with this reference:
\begin{equation}
    \begin{aligned}
    \rho^+(x_n) &= \tfrac{|\{x \in P : s_{\mathrm{ref}}(x) < s_{\Phi_k}(x_n)\}|}{|P|}, \\
    \tau^-(x_n) &= \tfrac{|\{x \in N : s_{\mathrm{ref}}(x) \geq s_{\Phi_k}(x_n)\}|}{|N|},
    \end{aligned}
    \label{eq:rank_context}
\end{equation}
For $x_n \in N$, we use $\rho^-(x_n) = |\{x \in N : s_{\mathrm{ref}}(x) < s_{\Phi_k}(x_n)\}|/|N|$ and $\tau^+(x_n) = |\{x \in P : s_{\mathrm{ref}}(x) \leq s_{\Phi_k}(x_n)\}|/|P|$. For either class, $\rho$ measures within-class position. The cross-class fraction $\tau$ counts reference negatives scoring at or above a positive, or reference positives scoring at or below a negative. This turns the bare correctness bit into a ranking-shaped critique aligned with the ranking objective. Templates, a worked $\mu_f$ example, and the per-signal ablation are in Appendix~\ref{sec:appendix_feedback_details}.
\begin{table*}[!b]
\centering
\small
\setlength{\tabcolsep}{5pt}

\caption{Test AUROC and balanced accuracy (\%) on three MIMIC disease tasks. Ranking-PE outperforms Accuracy-PE on AUROC across both backbones and all diseases. PE results are means; per-disease SDs appear in Appendix Table~\ref{tab:main_sd}. Within the Qwen3-VL-8B + SFT and MedGemma-4B blocks, bold and underline indicate the best and second-best values, respectively, including ties.}
\label{tab:main_results}
\resizebox{0.98\textwidth}{!}{\begin{tabular}{lcccccccc}
\toprule
 & \multicolumn{2}{c}{Atelectasis} & \multicolumn{2}{c}{Cardiomegaly} & \multicolumn{2}{c}{Consolidation} & \multicolumn{2}{c}{Avg} \\
\cmidrule(lr){2-3}\cmidrule(lr){4-5}\cmidrule(lr){6-7}\cmidrule(lr){8-9}
Method & AUROC & Bal.\ Acc. & AUROC & Bal.\ Acc. & AUROC & Bal.\ Acc. & AUROC & Bal.\ Acc. \\
\midrule
Qwen3-VL-8B & 65.4 & 64.0 & 61.3 & 51.2 & 74.0 & 64.4 & 66.9 & 59.9 \\
Gemma3-4B & 38.2 & 48.9 & 55.1 & 49.7 & 48.6 & 54.3 & 47.3 & 51.0 \\
GPT-4o-mini                                     & 60.1 & 60.2 & 54.7 & 50.5 & 70.5 & 53.0 & 61.8 & 54.6 \\
GPT-5-mini                                      & 50.6 & 58.2 & 71.5 & 66.8 & 75.9 & 64.6 & 66.0 & 63.2 \\
Gemini-2.5-flash                                & 46.5 & 54.3 & 69.0 & 58.3 & 75.1 & 66.4 & 63.5 & 59.7 \\
Claude-Sonnet-4.6                               & 51.4 & 59.3 & 64.2 & 61.4 & 74.0 & 67.4 & 63.2 & 62.7 \\
\midrule
Qwen3-VL-8B + SFT (VE-tuned) & 63.3 & 45.7 & 68.0 & 59.5 & 72.6 & 62.5 & 68.0 & 55.9 \\
\quad + Accuracy-PE & 63.6 & 45.4 & 68.1 & 58.0 & 72.4 & 61.8 & 68.0 & 55.1 \\
\quad + BAcc-Select & 63.9 & 43.7 & 67.6 & 60.6 & 71.4 & \underline{63.3} & 67.6 & 55.9 \\
\quad + Class-Weighted PE & 64.3 & 44.2 & 68.2 & 61.3 & 71.5 & 61.8 & 68.0 & 55.8 \\
\quad + Scalar-AUROC PE & \underline{66.5} & \underline{65.5} & \underline{69.3} & \underline{63.1} & \underline{72.9} & 61.2 & \underline{69.6} & \underline{63.3} \\
\quad + Ranking-PE (ours) & \textbf{70.7} & \textbf{70.9} & \textbf{71.3} & \textbf{66.5} & \textbf{79.3} & \textbf{68.5} & \textbf{73.8} & \textbf{68.6} \\
\midrule
MedGemma-4B & 46.7 & 47.5 & \underline{61.1} & 56.2 & 70.1 & 62.0 & \underline{59.3} & 55.2 \\
\quad + Accuracy-PE & 46.4 & 47.9 & 53.4 & 49.8 & 60.7 & 51.7 & 53.5 & 49.8 \\
\quad + BAcc-Select & \underline{49.5} & \underline{54.4} & 57.4 & \underline{57.5} & 70.2 & \textbf{64.0} & 59.0 & \underline{58.6} \\
\quad + Class-Weighted PE & 47.8 & 51.2 & 50.3 & 49.8 & 69.1 & 63.4 & 55.7 & 54.8 \\
\quad + Scalar-AUROC PE & 44.1 & 51.1 & 57.4 & 53.1 & \underline{71.3} & 60.3 & 57.6 & 54.8 \\
\quad + Ranking-PE (ours) & \textbf{69.1} & \textbf{60.5} & \textbf{63.0} & \textbf{57.6} & \textbf{77.1} & \underline{63.8} & \textbf{69.7} & \textbf{60.6} \\
\bottomrule

\end{tabular}}

\end{table*}

\subsubsection{Ranking-PE Pipeline}
\label{sec:pipeline_v2}

Ranking-PE composes three add-on alignments onto the Accuracy-PE base: Pair-Level Pareto Scoring, Ranking-Shaped Feedback, and AUROC-Based Selection (Figure~\ref{fig:method}).
\emph{(i)} \textbf{Pair-Level Pareto Scoring} (\S\ref{sec:pair_pareto_v2}): the row schema is replaced by the pair score $r(s(x^+),s(x^-))$, including half credit for ties, aligning the column average with empirical AUROC and Pareto dominance with pairwise ranking scores.
\emph{(ii)} \textbf{Ranking-Shaped Feedback} $\mu_f$ (\S\ref{sec:feedback_v2}): the reflector receives clinical-error-type, confidence-magnitude, and cross-example rank-context signals on top of the bare correctness flag.
\emph{(iii)} \textbf{AUROC-Based Selection} (Eq.~\ref{eq:final_selection}): after search, we select the candidate with the highest validation AUROC rather than accuracy:
\begin{equation}
    \Phi^* \;=\; \arg\max_{k}\ \mathrm{AUROC}(\Phi_k).
    \label{eq:final_selection}
\end{equation}

\section{Experiments}
\label{sec:experiments}

\subsection{Setup}
\label{sec:setup}

\noindent \textbf{Datasets.}
We use chest X-rays from MIMIC-CXR~\citep{johnson2019mimiccxr} linked to MIMIC-IV~\citep{johnson2023mimiciv} with structured EHR context (demographics, imaging view, laboratory values), following the input format $x = (\mathrm{image}, \mathrm{ehr})$ from \S\ref{sec:task_v2}. We evaluate three CheXpert competition pathologies~\citep{irvin2019chexpert} (Atelectasis, Cardiomegaly, and Consolidation), spanning class-imbalance regimes (overall cohort positive prevalence $96.2\%$ / $78.7\%$ / $68.9\%$, respectively) and dominant visual cues (subtle lung-field finding / global cardiac silhouette / diffuse opacity). Each disease is an independent binary task; dataset rationale and class-prevalence details are in Appendix~\ref{sec:appendix_impl_data}.

\noindent \textbf{Models.}
Our task-model backbones are open-source: Qwen3-VL-8B~\citep{bai2025qwen3} (8B general-purpose MLLM), MedGemma-4B~\citep{google2025medgemma} (medically pretrained), and Gemma3-4B~\citep{team2025gemma3} (non-medical matched-scale control to MedGemma). For zero-shot context, we additionally probe four closed-source models (GPT-4o-mini, GPT-5-mini, Gemini-2.5-flash, Claude-Sonnet-4.6) in Table~\ref{tab:main_results}. Stage~2 prompt evolution uses an LLM reflector; by default each task model serves as its own reflector, and we additionally evaluate two closed-source alternatives, GPT-5-nano and GPT-5-mini, on the Qwen3-VL-8B + SFT pipeline. The reflector consumes the same multimodal inputs as the task model but does not need logprob access; the open-source restriction on the task-model role is detailed in Appendix~\ref{sec:appendix_impl_mm}.

\noindent \textbf{Metrics.}
AUROC (Eq.~\ref{eq:pair_auroc_v2}) is our primary ranking metric. Balanced accuracy ($\tfrac{1}{2}(\mathrm{TPR}+\mathrm{TNR})$, the mean of sensitivity and specificity) at the default decoder threshold assesses whether ranking gains are accompanied by better binary diagnostic decisions under class imbalance. AUPRC is reported in Appendix Table~\ref{tab:main_full}.

\noindent \textbf{Multimodal interface.}
Reflective prompt-search frameworks~\citep{khattab2024dspy,agrawal2025gepa} are predominantly text-only; we extend them so that both the task model and the reflection LM operate on the same multimodal inputs (chest X-ray image, clinical profile, and lab context), grounding the reflector in visual evidence rather than a textual surrogate. Open-source backbones are served with logprob-enabled decoding so $s_\Phi(x)$ in Eq.~\ref{eq:score_v2} is read at no additional rollout cost; full serving and field-level details are in Appendix~\ref{sec:appendix_impl_mm}.

\noindent \textbf{Stage 1 and Stage 2 configurations.}
Stage~1 SFT runs with LLaMA-Factory~\citep{zheng2024llamafactory} and LoRA~\citep{hu2022lora} adapters; vision-encoder-tuned (VE-tuned) and vision-encoder-frozen (VE-frozen) variants differ only on whether the vision-tower parameters are trainable, and the final checkpoint is selected by held-out AUROC. Stage~2 maintains a population $\{\Phi_k\}_{k=1}^{K}$ and a Pareto-filtered candidate pool, evolved by reflective updates~\citep{agrawal2025gepa}. For each backbone, Accuracy-PE and Ranking-PE share the same task-model checkpoint, population size, rollout budget, and reflector, differing only on the three axes of \S\ref{sec:stage2}. We subsample uniformly to $\texttt{max\_pairs}=10{,}000$ when the pair count exceeds it (Table~\ref{tab:ablation_robustness}); full hyperparameters in Appendix~\ref{sec:appendix_impl}.

We also compare against balanced-accuracy selection, class-reweighted instance scores, and scalar-AUROC search. Their definitions are given in Appendix~\ref{sec:appendix_baselines}.

\begin{table*}[!b]
\centering
\small
\setlength{\tabcolsep}{8pt}

\caption{Test AUROC (\%) across model adaptation settings. Qwen3-VL-8B uses a $2\times2$ comparison of SFT and Ranking-PE, with vision-encoder tuning during SFT. Gemma3-4B and medically pretrained MedGemma-4B are evaluated before and after Ranking-PE. PE results are means; SDs appear in Appendix Table~\ref{tab:sft_results}. Bold indicates the best values within the Qwen3-VL-8B and MedGemma-4B blocks, including ties.}
\label{tab:foundations}
\resizebox{.9\textwidth}{!}{\begin{tabular}{llcccc}
\toprule
Model & Configuration & Atelectasis & Cardiomegaly & Consolidation & Avg \\
\midrule
Qwen3-VL-8B & Base & 65.4 & 61.3 & 74.0 & 66.9 \\
 & $+$ Ranking-PE & 63.5 & 65.8 & 76.0 & 68.4 \\
 & $+$ SFT & 63.3 & 68.0 & 72.6 & 68.0 \\
 & $+$ SFT $+$ Ranking-PE (ours) & \textbf{70.7} & \textbf{71.3} & \textbf{79.3} & \textbf{73.8} \\
\midrule
Gemma3-4B & Base & 38.2 & 55.1 & 48.6 & 47.3 \\
 & $+$ Ranking-PE & 50.5 & 55.1 & 49.3 & 51.6 \\
MedGemma-4B & Base (finetuned on clinical data) & 46.7 & 61.1 & 70.1 & 59.3 \\
 & $+$ Ranking-PE (ours)  & \textbf{69.1} & \textbf{63.0} & \textbf{77.1} & \textbf{69.7} \\
\bottomrule
\end{tabular}}

\end{table*}

\subsection{Results}
\label{sec:results}

\noindent \textbf{Main results~(Table~\ref{tab:main_results}).}
\label{sec:main_results}
Table~\ref{tab:main_results} reports three-run means for each PE method; per-disease SDs are provided in Appendix Table~\ref{tab:main_sd}. Ranking-PE yields substantial gains over Accuracy-PE on both main backbones. On Qwen3-VL-8B + SFT (VE-tuned), it reaches $73.8\%$ average AUROC and $68.6\%$ average balanced accuracy, improving over Accuracy-PE by $5.8$ and $13.5$ percentage points, respectively. On MedGemma-4B without SFT, it reaches $69.7\%$ average AUROC and $60.6\%$ average balanced accuracy, compared with $53.5\%$ and $49.8\%$ for Accuracy-PE. Accuracy-PE falls $5.8$ AUROC pp below the MedGemma base ($59.3\%$), while Ranking-PE improves on it by $10.4$ pp. Accuracy-targeted prompt evolution can therefore degrade the base model’s diagnostic ranking. Ranking-PE exceeds the corresponding base model on both metrics for each disease. On Atelectasis, Accuracy-PE remains close to the base on both backbones; at such extreme imbalance, validation accuracy provides little signal to distinguish candidate prompts.

Ranking-PE has higher AUROC than all three alternative-objective baselines on every disease for both backbones. The strongest of these baselines by average AUROC is scalar-AUROC search on Qwen ($69.6\%$) and balanced-accuracy selection on MedGemma ($59.0\%$), leaving margins of $4.2$ and $10.7$ percentage points. Ranking-PE also has the highest average balanced accuracy among the PE methods. These comparisons support guiding prompt evolution with ranking throughout the search, beyond changing the final selection metric or the aggregate objective alone.

\noindent \textbf{SFT staging~(Table~\ref{tab:foundations}; Appendix Table~\ref{tab:sft_results}).}
\label{sec:sft_foundation}
With Ranking-PE, average AUROC increases from $68.4\%$ without SFT to $70.3\%$ with VE-frozen SFT and $73.8\%$ with VE-tuned SFT. Vision-encoder tuning contributes $+3.5$ AUROC pp over frozen-encoder SFT, the largest single SFT-stage increment, with the largest per-disease gain on Atelectasis ($+4.8$ pp). Without SFT, Ranking-PE improves average AUROC by $1.5$ percentage points over the base model ($66.9\%$ to $68.4\%$). Prompt search benefits from visual adaptation, and tuning the vision encoder adds gains beyond language-side LoRA alone~\citep{li2023llavamed,he2024pefomed}.

\noindent \textbf{Medical pretraining~(Table~\ref{tab:foundations}).}
At matched parameter scale and architecture family, MedGemma-4B + Ranking-PE achieves $69.7\%$ average AUROC, compared with $51.6\%$ for Gemma3-4B + Ranking-PE, a gap of $18.1$ percentage points. Before prompt evolution, the corresponding averages are $59.3\%$ and $47.3\%$; Gemma3 is below chance on Atelectasis ($38.2\%$ AUROC). These results support medical adaptation as a foundation for prompt evolution. Medical pretraining provides another route to this foundation alongside vision-encoder-tuned SFT.

\noindent \textbf{Reflector substitution~(Table~\ref{tab:reflector}).}
Replacing the adapted Qwen self-reflector with GPT-5-nano or GPT-5-mini on the same SFT (VE-tuned) + Ranking-PE pipeline lowers average AUROC from $73.8\%$ to $66.7\%$ or $67.9\%$, respectively. Neither GPT reflector matches task-adapted self-reflection on any of the three diseases, in either AUROC or balanced accuracy. In this setting, a general-purpose reflector does not substitute for task-adapted self-reflection.

\noindent \textbf{Component ablation~(Table~\ref{tab:ablation}).}
\label{sec:ablation}
Changing only final selection from accuracy to AUROC reduces average test AUROC from $68.0\%$ to $61.3\%$. Adding the pair-level Pareto matrix raises it to $70.7\%$, exceeding Accuracy-PE by $2.7$ percentage points. Ranking-shaped feedback provides a further $3.1$ percentage points, reaching $73.8\%$. Average balanced accuracy increases from $62.7\%$ at the pair-matrix stage to $68.6\%$ with the full method. The pair-level representation contributes beyond ranking-shaped feedback and AUROC final selection: holding both fixed, replacing pair-level Pareto with an instance-level matrix lowers average AUROC from $73.8\%$ to $65.8\%$. Thus, the pair-level matrix provides gains before ranking feedback is added, while their combination gives the strongest results among these configurations. Changing final selection alone leaves candidate generation and parent selection guided by accuracy. Selecting by AUROC from this pool does not redirect the search toward ranking. Pair-level Pareto changes which candidates guide subsequent exploration, while ranking-shaped feedback helps the reflector propose ranking-oriented revisions. Per-disease SDs are reported in Appendix Table~\ref{tab:ablation_sd}.

\begin{table*}[t]
\centering
\small
\setlength{\tabcolsep}{4pt}

\caption{Component ablation on Qwen3-VL-8B + SFT (VE-tuned). Entries are mean test AUROC and balanced accuracy (\%); SDs appear in Appendix Table~\ref{tab:ablation_sd}. AUROC selection changes only Accuracy-PE's final selection criterion. The two component ablations replace pair-level Pareto with an instance-level matrix or ranking-shaped feedback with correctness-only feedback, respectively, while retaining AUROC final selection. Bold indicates the best value in each column.}
\label{tab:ablation}
\resizebox{0.98\textwidth}{!}{\begin{tabular}{lcccccccc}
\toprule
 & \multicolumn{2}{c}{Atelectasis} & \multicolumn{2}{c}{Cardiomegaly} & \multicolumn{2}{c}{Consolidation} & \multicolumn{2}{c}{Avg} \\
\cmidrule(lr){2-3}\cmidrule(lr){4-5}\cmidrule(lr){6-7}\cmidrule(lr){8-9}
Method & AUROC & Bal.\ Acc. & AUROC & Bal.\ Acc. & AUROC & Bal.\ Acc. & AUROC & Bal.\ Acc. \\
\midrule
Accuracy-PE & 63.6 & 45.4 & 68.1 & 58.0 & 72.4 & 61.8 & 68.0 & 55.1 \\
Accuracy-PE with AUROC selection & 43.7 & 51.5 & 68.3 & 56.5 & 72.0 & 63.9 & 61.3 & 57.3 \\
Ranking-PE (full) & \textbf{70.7} & \textbf{70.9} & \textbf{71.3} & \textbf{66.5} & \textbf{79.3} & \textbf{68.5} & \textbf{73.8} & \textbf{68.6} \\
\midrule
\multicolumn{9}{l}{\emph{Component ablations of Ranking-PE}} \\
\quad without pair-level Pareto & 56.0 & 51.1 & 68.2 & 61.4 & 73.3 & 60.0 & 65.8 & 57.5 \\
\quad without ranking-shaped feedback & 63.8 & 58.6 & 70.6 & 64.9 & 77.6 & 64.6 & 70.7 & 62.7 \\
\bottomrule
\vspace*{-7pt}
\end{tabular}}

\end{table*}

\noindent \textbf{Feedback-signal ablation~(Appendix Table~\ref{tab:ablation_feedback}).}
\label{sec:ablation_feedback}
On Consolidation, full ranking feedback achieves $79.3\%$ AUROC and $68.5\%$ balanced accuracy, compared with $77.6\%$ and $64.6\%$ for correctness-only feedback, gains of $1.7$ and $3.9$ percentage points. The clinical-error and confidence configurations yield AUROC values of $78.2\%$ and $79.8\%$, respectively. The reflector benefits from feedback beyond a bare correctness bit.

\noindent \textbf{Robustness~(Appendix Table~\ref{tab:ablation_robustness}).}
\label{sec:compute_cost}
On Cardiomegaly, the reported AUROC values range from $70.6\%$ to $71.3\%$ across pair caps of $300$, $1{,}000$, $3{,}000$, and $10{,}000$, a spread of $0.7$ percentage points. The small AUROC spread indicates that ranking performance is stable across the tested pair caps. Balanced accuracy ranges from $61.4\%$ to $67.3\%$. All reported configurations exceed Accuracy-PE on both metrics ($68.1\%$ AUROC and $58.0\%$ balanced accuracy).

\noindent \textbf{Qualitative analysis~(Appendix~\ref{sec:appendix_prompt_example}).}
Holding seed, budget, and reflector fixed, the two recipes evolve qualitatively different prompts. Accuracy-PE acquires decision-rule language: hard CT-ratio thresholds, asymmetric ``Yes''-pushing clauses, labs as overrides that flip the decision, and explicit normality criteria for answering ``No''. Ranking-PE acquires confidence-related guidance: overconfidence guards, guidance against missed findings, anti-dismissal clauses under image degradation, explicit limits on using labs alone, and a structured domain-knowledge etiology section. These differences illustrate how ranking-aware search changes the diagnostic guidance produced by the reflector, complementing the quantitative gains in the ablations. Side-by-side excerpts and full prompt text are in Appendix~\ref{sec:appendix_prompt_example} (Figures~\ref{fig:prompt_compare}, \ref{fig:prompt_example_cardiomegaly}).

\FloatBarrier
\section{Conclusion}
\label{sec:conclusion}

We argue that prompt evolution for MLLM-based clinical diagnosis should be guided by ranking rather than per-instance accuracy. We introduce \textbf{pair-level Pareto prompt evolution} (Ranking-PE), which redefines the optimizer's search event from instance correctness to positive-negative ordering to guide candidate comparison. Ranking-shaped reflector feedback guides prompt revision, while validation AUROC selects the final prompt. Combined with medical domain adaptation through vision-encoder-tuned SFT or medical pretraining, Ranking-PE outperforms Accuracy-PE on AUROC across two open-source MLLM families and three disease tasks on MIMIC, while also exceeding both Accuracy-PE and the corresponding base in average balanced accuracy. Component ablations show gains from the pair-level matrix and further improvements from ranking-shaped feedback.

The pair-level matrix preserves which positive-negative pairs each candidate correctly orders, allowing search to use information beyond a single aggregate metric; its column average equals empirical AUROC. Our formulation makes diagnostic ranking an explicit objective throughout reflective prompt evolution, guiding both candidate comparison and prompt revision without retraining, a surrogate loss, or additional rollouts.

\noindent \textbf{Limitations.}\label{sec:limitations}
Our evaluation focuses on MIMIC-IV multimodal clinical data across three diseases and two open-source MLLM families; evaluation on larger cohorts, across institutions, and on other imaging modalities remains future work. The pair-level Pareto formulation is presented for binary AUROC; multi-label and ordinal extensions are natural (one row schema per task or threshold) but untested. Evaluation uses small patient-disjoint test sets with limited representation of the negative minority. Closed-source baselines without top-$k$ logprobs fall back to a 3-level verbal score (Table~\ref{tab:main_results}); comparisons emphasize the open-source rows our recipe targets.

\section*{LLM Usage Statement}

\noindent \textbf{Generative AI as a component of the method.}
Large language models are not an incidental tool in this work but its object of
study and its machinery. Multimodal LLMs serve as the task models whose ranking
behaviour we measure and optimize, and an LLM reflector drives the
prompt-evolution search itself (\S\ref{sec:method}, \S\ref{sec:stage2}). The
task-model role is filled by Qwen3-VL-8B~\citep{bai2025qwen3},
MedGemma-4B~\citep{google2025medgemma} and Gemma3-4B~\citep{team2025gemma3}; the
reflector role is filled by the task model itself by default, and by GPT-5-nano
or GPT-5-mini in the reflector-substitution probe. Four closed-source models are
additionally probed zero-shot for context in Table~\ref{tab:main_results}. Roles,
serving configuration and hyperparameters are documented in \S\ref{sec:setup} and
Appendix~\ref{sec:appendix_impl_mm}--\ref{sec:appendix_impl_pe}; every reported
number is produced by the pipeline described in the paper.

\noindent \textbf{Generative AI in the preparation of this paper.}
We used OpenAI Codex to assist with language editing, wording suggestions, and to provide feedback on clarity and presentation. The authors reviewed and approved the resulting revisions.

We have reviewed all AI-assisted work. We take responsibility for the final content of this work, including text, claims or artifacts produced with the aid of generative AI.

{\small
\bibliographystyle{ieeenat_fullname}
\bibliography{refs}

@article{saito2015prplot,
  title={The precision-recall plot is more informative than the ROC plot when evaluating binary classifiers on imbalanced datasets},
  author={Saito, Takaya and Rehmsmeier, Marc},
  journal={{PLOS One}},
  volume={10},
  number={3},
  pages={e0118432},
  year={2015},
  publisher={Public Library of Science}
}

@inproceedings{davis2006prroc,
  title={The relationship between {Precision-Recall} and {ROC} curves},
  author={Davis, Jesse and Goadrich, Mark},
  booktitle={Proceedings of the 23rd international conference on Machine learning},
  pages={233--240},
  year={2006}
}

@book{pepe2003statistical,
  title={The statistical evaluation of medical tests for classification and prediction},
  author={Pepe, Margaret Sullivan},
  year={2003},
  publisher={Oxford university press}
}

@article{wynants2020covidreview,
  title={Prediction models for diagnosis and prognosis of {COVID-19}: systematic review and critical appraisal},
  author={Wynants, Laure and Van Calster, Ben and Collins, Gary S and Riley, Richard D and Heinze, Georg and Schuit, Ewoud and Albu, Elena and Arshi, Banafsheh and Bellou, Vanesa and Bonten, Marc MJ and others},
  journal={{BMJ}},
  volume={369},
  year={2020},
  publisher={British Medical Journal Publishing Group}
}

@article{collins2015tripod,
  title={Transparent reporting of a multivariable prediction model for individual prognosis or diagnosis ({TRIPOD}): the {TRIPOD} statement},
  author={Collins, Gary S and Reitsma, Johannes B and Altman, Douglas G and Moons, Karel GM},
  journal={Journal of British Surgery},
  volume={102},
  number={3},
  pages={148--158},
  year={2015},
  publisher={Oxford University Press}
}

@inproceedings{holtzman2021surfaceform,
  title={Surface form competition: Why the highest probability answer isn’t always right},
  author={Holtzman, Ari and West, Peter and Shwartz, Vered and Choi, Yejin and Zettlemoyer, Luke},
  booktitle={Proceedings of the 2021 Conference on Empirical Methods in Natural Language Processing},
  pages={7038--7051},
  year={2021}
}

@article{kadavath2022pknow,
  title={Language models (mostly) know what they know},
  author={Kadavath, Saurav and Conerly, Tom and Askell, Amanda and Henighan, Tom and Drain, Dawn and Perez, Ethan and Schiefer, Nicholas and Hatfield-Dodds, Zac and DasSarma, Nova and Tran-Johnson, Eli and others},
  journal={arXiv preprint arXiv:2207.05221},
  year={2022}
}

@inproceedings{li2023llavamed,
  title={{LLaVA-Med}: Training a large language-and-vision assistant for biomedicine in one day},
  author={Li, Chunyuan and Wong, Cliff and Zhang, Sheng and Usuyama, Naoto and Liu, Haotian and Yang, Jianwei and Naumann, Tristan and Poon, Hoifung and Gao, Jianfeng},
  booktitle={Advances in Neural Information Processing Systems},
  volume={36},
  pages={28541--28564},
  year={2023}
}

@article{tu2024medpalmm,
  title={Towards generalist biomedical {AI}},
  author={Tu, Tao and Azizi, Shekoofeh and Driess, Danny and Schaekermann, Mike and Amin, Mohamed and Chang, Pi-Chuan and Carroll, Andrew and Lau, Charles and Tanno, Ryutaro and Ktena, Ira and others},
  journal={{NEJM AI}},
  volume={1},
  number={3},
  pages={AIoa2300138},
  year={2024},
  publisher={Massachusetts Medical Society}
}

@article{zhang2024biomedgpt,
  title={A generalist vision--language foundation model for diverse biomedical tasks},
  author={Zhang, Kai and Zhou, Rong and Adhikarla, Eashan and Yan, Zhiling and Liu, Yixin and Yu, Jun and Liu, Zhengliang and Chen, Xun and Davison, Brian D and Ren, Hui and others},
  journal={Nature medicine},
  volume={30},
  number={11},
  pages={3129--3141},
  year={2024},
  publisher={Nature Publishing Group US New York}
}

@article{google2025medgemma,
  title={{MedGemma} technical report},
  author={Sellergren, Andrew and Kazemzadeh, Sahar and Jaroensri, Tiam and Kiraly, Atilla and Traverse, Madeleine and Kohlberger, Timo and Xu, Shawn and Jamil, Fayaz and Hughes, C{\'\i}an and Lau, Charles and others},
  journal={arXiv preprint arXiv:2507.05201},
  year={2025}
}

@article{bai2025qwen3,
  title={{Qwen3-VL} technical report},
  author={Bai, Shuai and Cai, Yuxuan and Chen, Ruizhe and Chen, Keqin and Chen, Xionghui and Cheng, Zesen and Deng, Lianghao and Ding, Wei and Gao, Chang and Ge, Chunjiang and others},
  journal={arXiv preprint arXiv:2511.21631},
  year={2025}
}

@inproceedings{hu2022lora,
  title={{LoRA}: Low-rank adaptation of large language models},
  author={Hu, Edward J and Shen, Yelong and Wallis, Phillip and Allen-Zhu, Zeyuan and Li, Yuanzhi and Wang, Shean and Wang, Liang and Chen, Weizhu and others},
  booktitle={International Conference on Learning Representations},
  year={2022}
}

@inproceedings{dettmers2023qlora,
  title={{QLoRA}: Efficient finetuning of quantized {LLMs}},
  author={Dettmers, Tim and Pagnoni, Artidoro and Holtzman, Ari and Zettlemoyer, Luke},
  booktitle={Advances in Neural Information Processing Systems},
  volume={36},
  pages={10088--10115},
  year={2023}
}

@article{agrawal2025gepa,
  title={{GEPA}: Reflective prompt evolution can outperform reinforcement learning},
  author={Agrawal, Lakshya A and Tan, Shangyin and Soylu, Dilara and Ziems, Noah and Khare, Rishi and Opsahl-Ong, Krista and Singhvi, Arnav and Shandilya, Herumb and Ryan, Michael J and Jiang, Meng and others},
  journal={arXiv preprint arXiv:2507.19457},
  year={2025}
}

@article{khattab2024dspy,
  title={{DSPy}: Compiling declarative language model calls into self-improving pipelines},
  author={Khattab, Omar and Singhvi, Arnav and Maheshwari, Paridhi and Zhang, Zhiyuan and Santhanam, Keshav and Vardhamanan, Sri and Haq, Saiful and Sharma, Ashutosh and Joshi, Thomas T and Moazam, Hanna and others},
  journal={arXiv preprint arXiv:2310.03714},
  year={2023}
}

@inproceedings{pryzant2023apo,
  title={Automatic prompt optimization with ``gradient descent'' and beam search},
  author={Pryzant, Reid and Iter, Dan and Li, Jerry and Lee, Yin and Zhu, Chenguang and Zeng, Michael},
  booktitle={Proceedings of the 2023 conference on empirical methods in natural language processing},
  pages={7957--7968},
  year={2023}
}

@article{fernando2023promptbreeder,
  title={{Promptbreeder}: Self-referential self-improvement via prompt evolution},
  author={Fernando, Chrisantha and Banarse, Dylan and Michalewski, Henryk and Osindero, Simon and Rockt{\"a}schel, Tim},
  journal={arXiv preprint arXiv:2309.16797},
  year={2023}
}

@inproceedings{burges2005ranknet,
  title={Learning to rank using gradient descent},
  author={Burges, Chris and Shaked, Tal and Renshaw, Erin and Lazier, Ari and Deeds, Matt and Hamilton, Nicole and Hullender, Greg},
  booktitle={Proceedings of the 22nd international conference on Machine learning},
  pages={89--96},
  year={2005}
}

@inproceedings{burges2006lambdarank,
  title={Learning to rank with nonsmooth cost functions},
  author={Burges, Christopher and Ragno, Robert and Le, Quoc},
  booktitle={Advances in Neural Information Processing Systems},
  volume={19},
  year={2006}
}

@inproceedings{yuan2021aucmax,
  title={Large-scale robust deep {AUC} maximization: A new surrogate loss and empirical studies on medical image classification},
  author={Yuan, Zhuoning and Yan, Yan and Sonka, Milan and Yang, Tianbao},
  booktitle={Proceedings of the IEEE/CVF international conference on computer vision},
  pages={3040--3049},
  year={2021}
}

@inproceedings{irvin2019chexpert,
  title={{CheXpert}: A large chest radiograph dataset with uncertainty labels and expert comparison},
  author={Irvin, Jeremy and Rajpurkar, Pranav and Ko, Michael and Yu, Yifan and Ciurea-Ilcus, Silviana and Chute, Chris and Marklund, Henrik and Haghgoo, Behzad and Ball, Robyn and Shpanskaya, Katie and others},
  booktitle={Proceedings of the AAAI conference on artificial intelligence},
  volume={33},
  number={01},
  pages={590--597},
  year={2019}
}

@article{johnson2023mimiciv,
  title={{MIMIC-IV}, a freely accessible electronic health record dataset},
  author={Johnson, Alistair E W and Bulgarelli, Lucas and Shen, Lu and Gayles, Alvin and Shammout, Ayad and Horng, Steven and Pollard, Tom J and Hao, Sicheng and Moody, Benjamin and Gow, Brian and Lehman, Li-wei H and Celi, Leo A and Mark, Roger G},
  journal={Scientific Data},
  volume={10},
  number={1},
  pages={1},
  year={2023},
  publisher={Nature Publishing Group}
}

@article{chambon2024chexpertplus,
  title={{CheXpert Plus}: Augmenting a Large Chest X-ray Dataset with Text Radiology Reports, Patient Demographics and Additional Image Formats},
  author={Chambon, Pierre and Delbrouck, Jean-Benoit and Sounack, Thomas and Huang, Shih-Cheng and Chen, Zhihong and Varma, Maya and Truong, Steven Q H and Chuong, Chu The and Langlotz, Curtis P},
  journal={arXiv preprint arXiv:2405.19538},
  year={2024}
}

@article{pal2025rexvqa,
  title={{ReXVQA}: A Large-scale Visual Question Answering Benchmark for Generalist Chest X-ray Understanding},
  author={Pal, Ankit and Lee, Jung-Oh and Zhang, Xiaoman and Sankarasubbu, Malaikannan and Roh, Seunghyeon and Kim, Won Jung and Lee, Meesun and Rajpurkar, Pranav},
  journal={arXiv preprint arXiv:2506.04353},
  year={2025}
}

@article{team2025gemma3,
  title={{Gemma 3} Technical Report},
  author={{Gemma Team}},
  journal={arXiv preprint arXiv:2503.19786},
  year={2025}
}

@inproceedings{zheng2024sglang,
  title={{SGLang}: Efficient execution of structured language model programs},
  author={Zheng, Lianmin and Yin, Liangsheng and Xie, Zhiqiang and Sun, Chuyue and Huang, Jeff and Yu, Cody Hao and Cao, Shiyi and Kozyrakis, Christos and Stoica, Ion and Gonzalez, Joseph E and Barrett, Clark and Sheng, Ying},
  booktitle={Advances in Neural Information Processing Systems},
  year={2024}
}

@inproceedings{zheng2024llamafactory,
  title={{LlamaFactory}: Unified Efficient Fine-Tuning of 100+ Language Models},
  author={Zheng, Yaowei and Zhang, Richong and Zhang, Junhao and Ye, Yanhan and Luo, Zheyan and Feng, Zhangchi and Ma, Yongqiang},
  booktitle={Proceedings of the 62nd Annual Meeting of the Association for Computational Linguistics (Volume 3: System Demonstrations)},
  year={2024}
}

@article{hanley1982auroc,
  title={The meaning and use of the area under a receiver operating characteristic ({ROC}) curve},
  author={Hanley, James A and McNeil, Barbara J},
  journal={Radiology},
  volume={143},
  number={1},
  pages={29--36},
  year={1982},
  publisher={Radiological Society of North America}
}

@article{he2024pefomed,
  title={{PeFoMed}: Parameter Efficient Fine-tuning of Multimodal Large Language Models for Medical Imaging},
  author={He, Jinlong and Li, Pengfei and Liu, Gang and He, Genrong and Chen, Zhaolin and Zhong, Shenjun},
  journal={arXiv preprint arXiv:2401.02797},
  year={2024}
}

@article{johnson2019mimiccxr,
  title={{MIMIC-CXR}, a de-identified publicly available database of chest radiographs with free-text reports},
  author={Johnson, Alistair E W and Pollard, Tom J and Berkowitz, Seth J and Greenbaum, Nathaniel R and Lungren, Matthew P and Deng, Chih-ying and Mark, Roger G and Horng, Steven},
  journal={Scientific Data},
  volume={6},
  number={1},
  pages={317},
  year={2019}
}

@article{nori2023medprompt,
  title={Can generalist foundation models outcompete special-purpose tuning? {C}ase study in medicine},
  author={Nori, Harsha and Lee, Yin Tat and Zhang, Sheng and Carignan, Dean and Edgar, Richard and Fusi, Nicolo and King, Nicholas and Larson, Jonathan and Li, Yuanzhi and Liu, Weishung and others},
  journal={arXiv preprint arXiv:2311.16452},
  year={2023}
}

@article{saporta2025symilemimic,
  author={Saporta, Adriel and Puli, Aahlad Manas and Goldstein, Mark and Ranganath, Rajesh},
  title={{Symile-MIMIC}: a multimodal clinical dataset of chest X-rays, electrocardiograms, and blood labs from {MIMIC-IV}},
  journal={PhysioNet},
  year={2025},
  note={Version 1.0.0},
  doi={10.13026/3vvj-s428}
}

@article{johnson2019mimiccxrjpg,
  author={Johnson, Alistair and Lungren, Matt and Peng, Yifan and Lu, Zhiyong and Mark, Roger and Berkowitz, Seth and Horng, Steven},
  title={{MIMIC-CXR-JPG}: chest radiographs with structured labels},
  journal={PhysioNet},
  year={2019},
  note={Version 2.0.0},
  doi={10.13026/8360-t248}
}
}
\clearpage
\appendix
\maketitlesupplementary
\section*{Appendix}

\noindent
This appendix supplements the main paper as follows. Appendix~\ref{sec:appendix_impl} collects implementation specifics deferred from \S\ref{sec:setup} (multimodal pipeline, SFT and prompt-evolution hyperparameters, and cohort/split construction). Appendix~\ref{sec:appendix_additional} reports additional results and ablations: SFT staging detail, the reflector-substitution probe, a cross-model prompt-transfer probe, and an AUPRC breakdown complementing Table~\ref{tab:main_results}. Appendix~\ref{sec:appendix_feedback_details} gives the per-example feedback rendering, a worked $\mu_f$ example, and the per-signal ablation. Appendix~\ref{sec:appendix_prompt_example} reproduces the seed, Accuracy-PE, and Ranking-PE prompts referenced in Figure~\ref{fig:prompt_compare} verbatim.

\section{Implementation details}
\label{sec:appendix_impl}

This appendix collects the implementation specifics deferred from \S\ref{sec:setup}. We organize them into four parts: the multimodal interface that bridges the task model and the reflection LM (\S\ref{sec:appendix_impl_mm}), Stage~1 SFT hyperparameters (\S\ref{sec:appendix_impl_sft}), Stage~2 prompt-evolution hyperparameters (\S\ref{sec:appendix_impl_pe}), and the cohort and split construction (\S\ref{sec:appendix_impl_data}).

\subsection{Multimodality Design and Implementation}
\label{sec:appendix_impl_mm}
The task module is implemented as a multimodal predictor whose signature exposes four input fields (\texttt{image}, \texttt{clinical\_profile}, \texttt{lab\_context}, \texttt{finding}) and two output fields (\texttt{answer}\,$\in$\,\{\texttt{Yes},\texttt{No}\}, \texttt{reasoning}). Image inputs are passed by reference (file path) and resolved at call time inside the model server, with a per-image pixel budget of $262{,}144$ pixels enforced upstream by the SFT preprocessor and inherited at inference time. The text trace is rendered with a $2048$-token cutoff.

The reflection step routes through a multimodal-aware instruction proposer that detects the presence of an image input on the reflective minibatch and renders each example with the image embedded as an inline \texttt{image\_url} block alongside the text trace (input fields, the candidate's prediction, and the per-example feedback string $\mu_f$). Without this routing, the reflection LM would receive only the textual surrogate and would lose access to the visual evidence that the task model saw.

We restrict the task-model role to open-source MLLMs for three reasons: (i) clinical inputs are sensitive and many deployments cannot route them to a closed endpoint; (ii) the SFT--prompt co-design recipe of \S\ref{sec:method} requires fine-tunable weights, which closed endpoints do not expose; (iii) the log-odds score $s_\Phi(x) = \log p(\mathtt{Yes}) - \log p(\mathtt{No})$ in Eq.~\ref{eq:score_v2} is read from top-$k$ logprobs at the answer-token position, which closed endpoints rarely expose.

Open-source backbones (Qwen3-VL-8B, MedGemma-4B, Gemma3-4B) are served with SGLang~\citep{zheng2024sglang} on four GPUs with tensor parallel size~$4$. Task models use greedy decoding (temperature $0$) and generate the answer before the reasoning. We extract logprobs at the answer-token position, so the score is not conditioned on the subsequently generated reasoning. Decoding requests use logprob-enabled chat completions ($\mathrm{top\_logprobs}=5$); the answer-token position is determined by the structured output field, and the log-odds score $s_\Phi(x)=\log p(\mathtt{Yes})-\log p(\mathtt{No})$ from Eq.~\ref{eq:score_v2} is read from a single decoding step. ``Yes'' and ``No'' are single-token labels in the tokenizers we use, so the score is determined in one decoding step; the emitted label is in the top-$k$ by construction, and the counterfactual label is in it whenever the model is not extremely confident. When the answer marker is not found, we assign a score of $1$ for a parsed ``Yes'' and $0$ for a parsed ``No'' ($0.035\%$ of test cases). Score extraction is read-only: the predicted label is the token the decoder emitted, independent of $s_\Phi$.

\subsection{Stage 1: SFT hyperparameters}
\label{sec:appendix_impl_sft}
Supervised fine-tuning is performed with LLaMA-Factory~\citep{zheng2024llamafactory}. We attach LoRA~\citep{hu2022lora} adapters of rank~$8$ to all linear projections of the language backbone (\texttt{lora\_target}=\texttt{all}) and use a shared training recipe across the SFT experiments, varying only the vision-tower flag between the VE-frozen and VE-tuned variants. Training uses AdamW with learning rate $2.0\times10^{-6}$, a cosine schedule with warmup ratio $0.03$, gradient clip $1.0$, and bf16 mixed precision. Per-device batch size is $2$ with gradient accumulation $1$; the run uses $5{,}000$ optimizer steps. Validation is run every $100$ steps with $\texttt{eval\_auroc}$ as the model-selection metric, and the final checkpoint is the best-AUROC step retained under \texttt{save\_total\_limit}=$5$. The VE-tuned variant additionally unfreezes the vision tower (\texttt{freeze\_vision\_tower}=\texttt{false}); the VE-frozen variant retains the default frozen vision tower. All other arguments default to the LLaMA-Factory shared configuration.

\subsection{Stage 2: prompt-evolution hyperparameters}
\label{sec:appendix_impl_pe}
Stage~2 follows the Pareto-based reflective search of \citet{agrawal2025gepa}. We use the framework's \texttt{auto}=\texttt{heavy} preset for all main-paper runs, which targets a candidate-prompt population of size $18$ (i.e., $K\leq 18$ in the notation of \S\ref{sec:pair_pareto_v2}); the framework then derives the total metric-call budget from the population size, the validation-set size, the reflection minibatch size, and the number of optimizable predictors. The reflection minibatch size is $3$ and we run $8$ concurrent inference threads. The reflection LM is by default the task model itself: Qwen3-VL-8B for the Qwen pipeline and MedGemma-4B for the MedGemma pipeline; the closed-source reflector probes (GPT-5-nano, GPT-5-mini) are evaluated only against the Qwen3-VL-8B + SFT pipeline. Final selection maximizes the validation column mean of the scores matrix: accuracy for Accuracy-PE and AUROC for Ranking-PE.

For the computational-cost analysis, we summarize results across the six model--disease settings. Estimated task-model evaluations average approximately $3{,}911$ per run for Accuracy-PE and $3{,}482$ for Ranking-PE. These estimates are derived from validation and minibatch evaluation counts and exclude reflection calls and test evaluation. Mean runtime for each method in each setting ranges from $39$ to $61$ minutes under a shared inference service, indicating comparable runtime scales. Pair scores reuse the predictions from each validation pass and require no additional task-model evaluations.

\begin{table}[H]
\centering
\small
\setlength{\tabcolsep}{10pt}
\renewcommand{\arraystretch}{1.1}
\caption{Sensitivity to the pair cap on Cardiomegaly with Qwen3-VL-8B + SFT (VE-tuned) and Ranking-PE. Values are test AUROC and balanced accuracy (\%).}
\label{tab:ablation_robustness}
\begin{tabular}{lcc}
\toprule
\texttt{max\_pairs} & AUROC & Bal.\ Acc. \\
\midrule
$300$                & $71.2$ & $67.3$ \\
$1{,}000$            & $70.6$ & $61.4$ \\
$3{,}000$            & $70.6$ & $65.4$ \\
$10{,}000$ (default) & $71.3$ & $66.5$ \\
\bottomrule
\end{tabular}
\end{table}

\paragraph{Alternative prompt-search objectives.}
\label{sec:appendix_baselines}
BAcc-Select retains unweighted instance-level correctness scores and binary correctness feedback, but selects the final prompt by validation balanced accuracy. Class-Weighted PE additionally weights each correctness score inversely to its class frequency on the validation set, so the mean score equals balanced accuracy. Scalar-AUROC PE assigns the same validation AUROC to every row in a candidate column, reducing Pareto comparisons to a scalar objective, and selects the final prompt by validation AUROC. All three baselines retain binary correctness feedback. Scalar-AUROC search differs from the final-selection-only ablation in Table~\ref{tab:ablation}, which retains instance-level correctness scores during search.

\paragraph{Pair-level Pareto.} The matrix $\tilde{M}$ contains pair scores $r(s_{\Phi_k}(x_i^+),s_{\Phi_k}(x_j^-))$, with half credit for ties (\S\ref{sec:pair_pareto_v2}). These scores are computed from the per-instance scores $\{s_{\Phi_k}(x)\}$ that the validation pass already produces. We use all positive--negative pairs when $|P|\cdot|N|\leq\texttt{max\_pairs}$, with a default cap of $10{,}000$. Otherwise, we sample $\texttt{max\_pairs}$ pairs uniformly without replacement for each run. The selected pair set is shared by all candidates and reused across iterations within the run. Table~\ref{tab:ablation_robustness} reports results on Cardiomegaly for $\texttt{max\_pairs}\in\{300,1{,}000,3{,}000,10{,}000\}$.

\paragraph{Ranking-shaped feedback.} The per-example feedback string $\mu_f$ (\S\ref{sec:feedback_v2}) concatenates the three signals in a fixed order: clinical error type, confidence magnitude, and cross-example rank context $(\rho^+,\tau^-)$. Concrete templates, a worked example, and the per-signal ablation are in Appendix~\ref{sec:appendix_feedback_details}.

\begin{table*}[!t]
\centering
\small
\setlength{\tabcolsep}{4pt}
\caption{Variability of the PE results in Table~\ref{tab:main_results}. Per-disease test AUROC and balanced accuracy (\%) are mean $\pm$ SD. Avg is the unweighted average of the reported disease means; no aggregate SD is reported.}
\label{tab:main_sd}
\resizebox{0.98\textwidth}{!}{\begin{tabular}{lcccccccc}
\toprule
 & \multicolumn{2}{c}{Atelectasis} & \multicolumn{2}{c}{Cardiomegaly} & \multicolumn{2}{c}{Consolidation} & \multicolumn{2}{c}{Avg} \\
\cmidrule(lr){2-3}\cmidrule(lr){4-5}\cmidrule(lr){6-7}\cmidrule(lr){8-9}
Method & AUROC & Bal.\ Acc. & AUROC & Bal.\ Acc. & AUROC & Bal.\ Acc. & AUROC & Bal.\ Acc. \\
\midrule
\multicolumn{9}{l}{Qwen3-VL-8B + SFT (VE-tuned)} \\
Accuracy-PE & $63.6\pm0.5$ & $45.4\pm0.6$ & $68.1\pm1.2$ & $58.0\pm1.4$ & $72.4\pm0.2$ & $61.8\pm1.0$ & 68.0 & 55.1 \\
BAcc-Select & $63.9\pm0.4$ & $43.7\pm0.4$ & $67.6\pm1.0$ & $60.6\pm0.7$ & $71.4\pm1.3$ & $63.3\pm0.6$ & 67.6 & 55.9 \\
Class-Weighted PE & $64.3\pm0.2$ & $44.2\pm0.4$ & $68.2\pm1.7$ & $61.3\pm1.4$ & $71.5\pm0.7$ & $61.8\pm3.2$ & 68.0 & 55.8 \\
Scalar-AUROC PE & $66.5\pm3.6$ & $65.5\pm0.8$ & $69.3\pm1.6$ & $63.1\pm1.1$ & $72.9\pm1.6$ & $61.2\pm3.9$ & 69.6 & 63.3 \\
Ranking-PE & $70.7\pm0.9$ & $70.9\pm4.3$ & $71.3\pm1.0$ & $66.5\pm2.1$ & $79.3\pm1.2$ & $68.5\pm1.2$ & 73.8 & 68.6 \\
\midrule
\multicolumn{9}{l}{MedGemma-4B} \\
Accuracy-PE & $46.4\pm0.4$ & $47.9\pm0.6$ & $53.4\pm0.9$ & $49.8\pm0.2$ & $60.7\pm2.6$ & $51.7\pm1.5$ & 53.5 & 49.8 \\
BAcc-Select & $49.5\pm2.3$ & $54.4\pm2.9$ & $57.4\pm2.2$ & $57.5\pm1.8$ & $70.2\pm1.6$ & $64.0\pm2.4$ & 59.0 & 58.6 \\
Class-Weighted PE & $47.8\pm1.6$ & $51.2\pm3.3$ & $50.3\pm4.8$ & $49.8\pm2.1$ & $69.1\pm1.5$ & $63.4\pm1.0$ & 55.7 & 54.8 \\
Scalar-AUROC PE & $44.1\pm2.5$ & $51.1\pm4.2$ & $57.4\pm1.9$ & $53.1\pm3.1$ & $71.3\pm1.1$ & $60.3\pm2.9$ & 57.6 & 54.8 \\
Ranking-PE & $69.1\pm3.1$ & $60.5\pm3.1$ & $63.0\pm4.5$ & $57.6\pm1.9$ & $77.1\pm3.8$ & $63.8\pm2.3$ & 69.7 & 60.6 \\
\bottomrule
\end{tabular}}
\end{table*}

\begin{table*}[!b]
\centering
\small
\setlength{\tabcolsep}{4pt}
\caption{Variability of the component ablation in Table~\ref{tab:ablation}. Per-disease AUROC and balanced accuracy (\%) are mean $\pm$ SD. Avg averages the disease means. Component ablations use the same replacements as in Table~\ref{tab:ablation}.}
\label{tab:ablation_sd}
\resizebox{0.98\textwidth}{!}{\begin{tabular}{lcccccccc}
\toprule
 & \multicolumn{2}{c}{Atelectasis} & \multicolumn{2}{c}{Cardiomegaly} & \multicolumn{2}{c}{Consolidation} & \multicolumn{2}{c}{Avg} \\
\cmidrule(lr){2-3}\cmidrule(lr){4-5}\cmidrule(lr){6-7}\cmidrule(lr){8-9}
Method & AUROC & Bal.\ Acc. & AUROC & Bal.\ Acc. & AUROC & Bal.\ Acc. & AUROC & Bal.\ Acc. \\
\midrule
Accuracy-PE & $63.6\pm0.5$ & $45.4\pm0.6$ & $68.1\pm1.2$ & $58.0\pm1.4$ & $72.4\pm0.2$ & $61.8\pm1.0$ & 68.0 & 55.1 \\
Accuracy-PE with AUROC selection & $43.7\pm2.8$ & $51.5\pm0.7$ & $68.3\pm1.9$ & $56.5\pm1.7$ & $72.0\pm0.0$ & $63.9\pm2.6$ & 61.3 & 57.3 \\
Ranking-PE (full) & $70.7\pm0.9$ & $70.9\pm4.3$ & $71.3\pm1.0$ & $66.5\pm2.1$ & $79.3\pm1.2$ & $68.5\pm1.2$ & 73.8 & 68.6 \\
\midrule
\multicolumn{9}{l}{\emph{Component ablations of Ranking-PE}} \\
\quad without pair-level Pareto & $56.0\pm4.4$ & $51.1\pm1.8$ & $68.2\pm0.1$ & $61.4\pm0.4$ & $73.3\pm0.5$ & $60.0\pm4.3$ & 65.8 & 57.5 \\
\quad without ranking-shaped feedback & $63.8\pm3.4$ & $58.6\pm2.1$ & $70.6\pm0.9$ & $64.9\pm0.4$ & $77.6\pm2.7$ & $64.6\pm2.1$ & 70.7 & 62.7 \\
\bottomrule
\end{tabular}}
\end{table*}

\subsection{Data construction}
\label{sec:appendix_impl_data}
We use the admission-level cohort from Symile-MIMIC~\citep{saporta2025symilemimic}, which links MIMIC-CXR images to MIMIC-IV records. Each admission contributes the earliest anteroposterior (AP) or posteroanterior (PA) chest X-ray acquired between 24 and 72 hours after admission. Disease labels are taken from the official MIMIC-CXR-JPG CheXpert labels~\citep{johnson2019mimiccxrjpg}, automatically extracted from radiology reports and matched to the selected image by \texttt{study\_id}. For each disease, we map labels $1$ and $0$ to $\mathtt{Yes}$ and $\mathtt{No}$, respectively, and exclude uncertain ($-1$) and missing labels. This filtering is applied independently for each disease, so exclusion from one task does not exclude an admission from the others. Each retained example is a tuple $(x_n,d_n,y_n)$ with $x_n=(\mathrm{image},\mathrm{ehr})$ and $y_n\in\{\mathtt{Yes},\mathtt{No}\}$ for a target disease $d_n$.

The clinical-profile field is composed of patient demographics, admission type, and imaging view; the lab-context field concatenates a fixed panel of MIMIC-IV laboratory values rendered as \texttt{Item:value} tokens, with missing entries written as \texttt{NA}. The three diseases reported in the headline tables (Atelectasis, Cardiomegaly, and Consolidation) are selected on three grounds. (i) \emph{Benchmark precedent}: all three appear among the competition pathologies adjudicated on the original CheXpert benchmark~\citep{irvin2019chexpert} and recur across recent CXR multimodal LLM evaluations~\citep{chambon2024chexpertplus,pal2025rexvqa}, providing external comparability with the broader medical-imaging literature. (ii) \emph{Imbalance-regime spread}: positive prevalence over the retained cohort (train, validation, and test combined) is $96.2\%$ for Atelectasis, $78.7\%$ for Cardiomegaly, and $68.9\%$ for Consolidation, with disease presence (``Yes'') as the positive class, so the same recipe is exercised across distinct imbalance regimes that our method is designed to address. (iii) \emph{Radiological-evidence diversity}: the three diseases differ in their dominant visual cue: a subtle local lung-field finding (Atelectasis), a global cardiac-silhouette / cardiothoracic-ratio measurement (Cardiomegaly), and a diffuse opacity finding (Consolidation). Thus, improvements that hold across all three are unlikely to reflect a single image cue.

The corresponding test-set positive prevalences are $96.6\%$, $74.2\%$, and $56.9\%$, respectively. We retain the Symile-MIMIC splits, with no patient overlap between training, validation, and test sets. After label filtering, the training/validation/test sets contain $901/230/146$ examples for Atelectasis, $4{,}232/349/194$ for Cardiomegaly, and $1{,}071/82/51$ for Consolidation. In Stage~2, we shuffle the training pool using the run seed and take $500$ examples (\texttt{max\_train}). For the main experiments, Stage~1 checkpoint selection and Stage~2 prompt selection use the same validation split where both stages are run. We use the full validation split for selection and the full test split only for final evaluation.

\makeatletter
\setlength{\@dblfpsep}{12pt plus 2fil}
\makeatother

\section{Additional results and ablations}
\label{sec:appendix_additional}

The main paper carries the foundation argument with Table~\ref{tab:foundations} (\S\ref{sec:sft_foundation}). Table~\ref{tab:sft_results} reproduces the underlying SFT-stage detail for completeness. Table~\ref{tab:reflector} expands the reflector substitution probe cited in the main paper's foundation analysis. Table~\ref{tab:transfer_results} reports a cross-model prompt-transfer probe.

\subsection{Variation across runs}
\label{sec:appendix_variation}

Table~\ref{tab:main_sd} reports test performance (mean $\pm$ SD) across three prompt-evolution runs per setting. The three runs reuse the same task-model checkpoint without repeating SFT.

\begin{samepage}
\subsection{SFT staging detail}
Table~\ref{tab:sft_results} compares SFT settings under Ranking-PE; vision-encoder tuning yields the best AUROC on all three Qwen3-VL-8B tasks.
\par
\end{samepage}

\subsection{Reflector substitution}
Table~\ref{tab:reflector} compares self-reflection with GPT-5-nano and GPT-5-mini reflectors for Qwen3-VL-8B + SFT.

\subsection{Cross-model prompt transfer}
Table~\ref{tab:transfer_results} tests MedGemma-4B prompts on Qwen3-VL-8B + SFT without further prompt optimization.

\subsection{AUPRC breakdown}
\label{sec:appendix_full}
Table~\ref{tab:main_full} gives complementary AUPRC results~\citep{saito2015prplot,davis2006prroc}.

\begin{table*}[!tp]
\centering
\small
\setlength{\tabcolsep}{3pt}
\renewcommand{\arraystretch}{1.12}
\caption{Test AUROC (\%) across SFT settings and model backbones. All configurations use Ranking-PE. On Qwen, average AUROC increases with each SFT stage, with vision-encoder tuning providing the largest increment. Disease entries report mean $\pm$ SD; Avg is the unweighted average of the reported disease means.}
\label{tab:sft_results}
\begin{tabular*}{\textwidth}{@{\extracolsep{\fill}}llcccc@{}}
\toprule
Model & SFT setting & Atelectasis & Cardiomegaly & Consolidation & Avg \\
\midrule
Qwen3-VL-8B & None & $63.5\pm3.6$ & $65.8\pm0.4$ & $76.0\pm3.8$ & 68.4 \\
 & VE-frozen & $65.9\pm0.8$ & $67.3\pm1.2$ & $77.6\pm0.8$ & 70.3 \\
 & VE-tuned & $70.7\pm0.9$ & $71.3\pm1.0$ & $79.3\pm1.2$ & 73.8 \\
\midrule
Gemma3-4B & None & $50.5\pm4.3$ & $55.1\pm0.0$ & $49.3\pm1.3$ & 51.6 \\
\addlinespace[2pt]
MedGemma-4B & None & $69.1\pm3.1$ & $63.0\pm4.5$ & $77.1\pm3.8$ & 69.7 \\
\bottomrule
\end{tabular*}
\end{table*}


\begin{table*}[!tp]
\centering
\small
\setlength{\tabcolsep}{4pt}
\caption{Reflector substitution on Qwen3-VL-8B + SFT (VE-tuned). Self-reflection outperforms both GPT reflectors on all three diseases in AUROC and balanced accuracy. Values are test AUROC and balanced accuracy (\%). Avg is the unweighted average across diseases.}
\label{tab:reflector}
\resizebox{0.98\textwidth}{!}{\begin{tabular}{lcccccccc}
\toprule
 & \multicolumn{2}{c}{Atelectasis} & \multicolumn{2}{c}{Cardiomegaly} & \multicolumn{2}{c}{Consolidation} & \multicolumn{2}{c}{Avg} \\
\cmidrule(lr){2-3}\cmidrule(lr){4-5}\cmidrule(lr){6-7}\cmidrule(lr){8-9}
Method & AUROC & Bal.\ Acc. & AUROC & Bal.\ Acc. & AUROC & Bal.\ Acc. & AUROC & Bal.\ Acc. \\
\midrule
Self-reflector & 70.7 & 70.9 & 71.3 & 66.5 & 79.3 & 68.5 & 73.8 & 68.6 \\
GPT-5-nano reflector & 59.9 & 55.7 & 68.0 & 61.8 & 72.1 & 57.5 & 66.7 & 58.3 \\
GPT-5-mini reflector & 58.3 & 50.0 & 68.0 & 61.8 & 77.3 & 58.6 & 67.9 & 56.8 \\
\bottomrule
\end{tabular}}
\end{table*}

\begin{table*}[!tp]
\centering
\small
\setlength{\tabcolsep}{5pt}
\caption{Cross-model prompt transfer from MedGemma-4B to Qwen3-VL-8B + SFT without further prompt optimization. On average, the transferred prompt outperforms Accuracy-PE run on the target model, suggesting that the discovered instructions are not entirely model-specific. Values are test AUROC and balanced accuracy (\%); Avg is the unweighted average across diseases.}
\label{tab:transfer_results}
\resizebox{\textwidth}{!}{\begin{tabular}{lcccccccc}
\toprule
 & \multicolumn{2}{c}{Atelectasis} & \multicolumn{2}{c}{Cardiomegaly} & \multicolumn{2}{c}{Consolidation} & \multicolumn{2}{c}{Avg} \\
\cmidrule(lr){2-3}\cmidrule(lr){4-5}\cmidrule(lr){6-7}\cmidrule(lr){8-9}
Method & AUROC & Bal.\ Acc. & AUROC & Bal.\ Acc. & AUROC & Bal.\ Acc. & AUROC & Bal.\ Acc. \\
\midrule
Qwen3-VL-8B + SFT (VE-tuned) & 63.3 & 45.7 & 68.0 & 59.5 & 72.6 & 62.5 & 68.0 & 55.9 \\
\quad + Accuracy-PE & 63.6 & 45.4 & 68.1 & 58.0 & 72.4 & 61.8 & 68.0 & 55.1 \\
\quad + Ranking-PE (ours) & 70.7 & 70.9 & 71.3 & 66.5 & 79.3 & 68.5 & 73.8 & 68.6 \\
\cmidrule(lr){1-9}
\multicolumn{9}{l}{\textit{Cross-model transfer: prompt optimized on MedGemma-4B, evaluated on Qwen3-VL-8B + SFT}} \\
\quad + Transferred Ranking-PE prompt            & 64.4 & 58.2 & 68.1 & 63.5 & 78.1 & 66.1 & 70.2 & 62.6 \\
\bottomrule
\end{tabular}}
\par\medskip
\centering
\small
\setlength{\tabcolsep}{5pt}
\caption{Test AUPRC (\%). For each disease, PE methods report mean $\pm$ SD; models without PE report means only. Avg is the unweighted average of the reported disease means.}
\label{tab:main_full}
\begin{tabular}{lcccc}
\toprule
Method & Atelectasis & Cardiomegaly & Consolidation & Avg \\
\midrule
Qwen3-VL-8B & $97.9$ & $82.9$ & $79.0$ & $86.6$ \\
\midrule
Qwen3-VL-8B + SFT (VE-tuned) & $98.2$ & $85.4$ & $81.4$ & $88.3$ \\
\quad + Accuracy-PE & $98.2 \pm 0.1$ & $84.0 \pm 1.3$ & $81.3 \pm 0.2$ & $87.8$ \\
\quad + BAcc-Select & $98.2\pm0.1$ & $84.0\pm2.8$ & $76.7\pm4.3$ & 86.3 \\
\quad + Class-Weighted PE & $98.3\pm0.0$ & $84.8\pm1.5$ & $79.5\pm1.6$ & 87.5 \\
\quad + Scalar-AUROC PE & $98.1\pm0.2$ & $84.0\pm1.3$ & $79.2\pm1.1$ & 87.1 \\
\quad + Ranking-PE & $98.4 \pm 0.2$ & $86.1 \pm 1.6$ & $84.2 \pm 0.9$ & $89.6$ \\
\midrule
MedGemma-4B & $94.8$ & $81.5$ & $78.4$ & $84.9$ \\
\quad + Accuracy-PE & $95.4 \pm 1.0$ & $77.0 \pm 0.2$ & $68.2 \pm 1.3$ & $80.2$ \\
\quad + BAcc-Select & $96.7\pm0.2$ & $78.8\pm1.0$ & $79.9\pm1.3$ & 85.1 \\
\quad + Class-Weighted PE & $96.9\pm0.7$ & $74.7\pm3.5$ & $77.1\pm0.8$ & 82.9 \\
\quad + Scalar-AUROC PE & $96.8\pm0.2$ & $78.3\pm1.6$ & $81.0\pm2.3$ & 85.4 \\
\quad + Ranking-PE & $98.6\pm0.3$ & $81.2\pm3.1$ & $83.5\pm3.7$ & $87.8$ \\
\bottomrule
\end{tabular}

\end{table*}

\clearpage
\begin{figure*}[p]
\begin{minipage}{\textwidth}
\setlength{\columnwidth}{\dimexpr(\textwidth-\columnsep)/2\relax}
\addtolength{\abovedisplayskip}{-2pt}
\addtolength{\belowdisplayskip}{-2pt}
\addtolength{\abovedisplayshortskip}{-2pt}
\addtolength{\belowdisplayshortskip}{-2pt}
\section{Ranking-shaped feedback details}
\label{sec:appendix_feedback_details}
\noindent
\begin{minipage}[t]{\columnwidth}
\vspace{0pt}
The per-example feedback $\mu_f$ described in \S\ref{sec:feedback_v2} is assembled by newline-concatenating three signals into a single string the reflector reads verbatim. We give the concrete rendering of each signal, a worked $\mu_f$ example, and the per-signal ablation here.

\paragraph{Clinical error type.}
The error-type signal replaces the bare correctness bit with one of $e \in \{\mathrm{TP}, \mathrm{TN}, \mathrm{FN}, \mathrm{FP}\}$ (confusion-matrix labels with $\mathtt{Yes}$ as positive), rendered as a full-sentence string per Table~\ref{tab:error_type_rendering}. Only $e = \mathrm{FN}$ carries the ``(clinically dangerous)'' annotation, reflecting the asymmetric error costs in screening, where a missed positive is materially worse than a false alarm. Both error cases (FN, FP) carry a short tactical nudge that correct cases (TP, TN) do not.

{\centering
\small
\captionof{table}{Exact feedback string appended to $\mu_f$ for each value of
$e(\hat{y}_{n,k}, y_n)$. Here $d$ is the target disease for the instance
(\S\ref{sec:task_v2}). Only $e = \mathrm{FN}$ carries the
``(clinically dangerous)'' annotation.}
\label{tab:error_type_rendering}
\begin{tabular}{@{}lp{0.78\linewidth}@{}}
\toprule
$e$ & String appended to $\mu_f$ \\
\midrule
$\mathrm{TP}$ & \texttt{TRUE POSITIVE --- correctly identified $d$ as present.} \\
$\mathrm{TN}$ & \texttt{TRUE NEGATIVE --- correctly ruled out $d$.} \\
$\mathrm{FN}$ & \texttt{FALSE NEGATIVE (clinically dangerous) --- missed $d$. Re-examine the image for subtle evidence.} \\
$\mathrm{FP}$ & \texttt{FALSE POSITIVE --- over-called $d$. Require converging evidence before committing.} \\
\bottomrule
\end{tabular}
\par\medskip\par}

\textbf{Per-signal ablation.} Table~\ref{tab:ablation_feedback} compares cumulative feedback configurations on Consolidation with pair-level Pareto scoring and validation AUROC selection held fixed. Basic corresponds to the ``without ranking-shaped feedback'' row in Table~\ref{tab:ablation}, and full corresponds to the ``Ranking-PE (full)'' row. Full feedback improves AUROC by $1.7$ percentage points and balanced accuracy by $3.9$ percentage points over Basic.
\begin{table}[H]
\centering
\small
\setlength{\tabcolsep}{4pt}
\renewcommand{\arraystretch}{1.1}
\caption{Feedback-signal ablation on Consolidation with Qwen3-VL-8B + SFT (VE-tuned). Signals are added cumulatively with pair-level Pareto and AUROC final selection. Values are test AUROC and balanced accuracy (\%).}
\label{tab:ablation_feedback}
\begin{tabular}{lcc}
\toprule
Feedback level & AUROC & Bal.\ Acc. \\
\midrule
Basic (correctness only)                  & 77.6 & 64.6 \\
\quad + clinical error type               & 78.2 & 65.5 \\
\quad + confidence $|s(x)|$               & 79.8 & 63.8 \\
\quad + cross-example rank context (full) & 79.3 & 68.5 \\
\bottomrule
\end{tabular}
\end{table}

\end{minipage}\hfill
\begin{minipage}[t]{\columnwidth}
\vspace{0pt}
\paragraph{Confidence magnitude.}
The signed log-odds $s_{\Phi_k}(x_n)$ is rendered alongside a plain-language bucket
\begin{equation}
    \begin{aligned}
    b(|s|) = {}&\text{low}\ (|s| < 1), \\
    &\text{moderate}\ (1 \le |s| < 3), \\
    &\text{high}\ (3 \le |s| < 6), \\
    &\text{extremely high}\ (|s| \ge 6),
    \end{aligned}
    \label{eq:conf_bucket}
\end{equation}
with thresholds fixed to the magnitude scale of $s_\Phi$ observed in our experiments. An overconfidence flag fires when $|s| \geq 3$ and the prediction is incorrect.

\paragraph{Cross-example rank context.}
The $(\rho^+, \tau^-)$ pair defined in Eq.~\ref{eq:rank_context} is computed from cached per-instance validation scores $s_{\mathrm{ref}}$ over $P \cup N$ from the most recently fully evaluated candidate. It can be used independently of the pair-level Pareto matrix. $\mu_f$ is still called per-example: inputs are the single-instance triple $(x_n, \hat{y}_{n,k}, y_n)$, and the injected rank context is a pair of scalar summaries computed against the reference validation distribution, not an enumeration over other examples. Cohort-level pattern discovery (e.g., the prompt is systematically overconfident on portable AP views) is delegated to the reflector, which reads the full minibatch of $\mu_f$ outputs together. The optimizer's interface is therefore unchanged.

\paragraph{Worked example.}
Figure~\ref{fig:feedback_templates} shows a complete $\mu_f$ output assembled from all three signals on a false-negative positive example.

{\centering\small
\fbox{\begin{minipage}{0.93\linewidth}
\raggedright
\textbf{Complete $\mu_f$ output for a false-negative positive example
($e = \mathrm{FN}$, $x_n \in P$).}\\[0.4em]
\texttt{\textcolor{red!70!black}{FALSE NEGATIVE (clinically dangerous) --- missed $d$. Re-examine the image for subtle evidence.}\\
Model confidence: s(x) = \textcolor{blue}{-1.50}\ (\textcolor{purple}{moderate}).\\
Val ranking: this +example s=\textcolor{blue}{-1.50} ranks
\textcolor{orange}{3/40} within val positives
(\textcolor{orange}{5\%} below it); \textcolor{red}{75\%} of val
negatives scored at or above it.}
\end{minipage}}
\captionof{figure}{Full $\mu_f$ string appended for one example (illustrative values), assembled by
newline-concatenating the three signals.
Line~1 is the clinical error type (rendered per
Table~\ref{tab:error_type_rendering} for $e = \mathrm{FN}$);
line~2 is the confidence magnitude (Eq.~\ref{eq:conf_bucket});
line~3 is the cross-example rank context (Eq.~\ref{eq:rank_context}).
Color-coded dynamic regions:
\textcolor{red!70!black}{rendered error-type outcome};
\textcolor{blue}{signed score $s_{\Phi_k}(x_n)$};
\textcolor{purple}{magnitude bucket $b(|s|)$};
\textcolor{orange}{within-class rank position and $\rho^+(x_n)$};
\textcolor{red}{cross-class threat fraction $\tau^-(x_n)$}.
The low $\rho^+(x_n) = 5\%$ together with a high $\tau^-(x_n) = 75\%$
show that this missed positive ranks low relative to the reference positives
and below or tied with many reference negatives.
For a negative example $x_n \in N$, the rank-context line swaps
$P \leftrightarrow N$ and reports $(\rho^-, \tau^+)$ as defined in
\S\ref{sec:feedback_v2} (the fraction of reference negative scores below its current score and
the fraction of reference positive scores at or below it) and replaces
``+example'' with ``-example''.}
\label{fig:feedback_templates}
\par}
\end{minipage}

\end{minipage}
\end{figure*}

%
%

\definecolor{aceFill}{RGB}{253,237,237}
\definecolor{aceEdge}{RGB}{200,80,80}
\definecolor{rpeFill}{RGB}{235,247,237}
\definecolor{rpeEdge}{RGB}{60,140,80}
\definecolor{quoteGray}{RGB}{75,75,75}

\newtcolorbox{aceCard}[1][]{
  enhanced, colback=aceFill, colframe=aceEdge, boxrule=0.4pt,
  arc=2pt, left=5pt, right=5pt, top=4pt, bottom=4pt,
  fontupper=\footnotesize, before skip=4pt, after skip=2pt, #1
}
\newtcolorbox{rpeCard}[1][]{
  enhanced, colback=rpeFill, colframe=rpeEdge, boxrule=0.4pt,
  arc=2pt, left=5pt, right=5pt, top=4pt, bottom=4pt,
  fontupper=\footnotesize, before skip=4pt, after skip=2pt, #1
}

\providecommand{\faTimes}{\ensuremath{\boldsymbol{\times}}}
\providecommand{\faCheck}{\ensuremath{\boldsymbol{\checkmark}}}
\newcommand{\promptquote}[1]{%
  {\color{quoteGray}\itshape\footnotesize ``#1''}\par%
}
\newcommand{\flawTag}[1]{{\color{aceEdge}\bfseries\footnotesize\faTimes\ #1}\par}
\newcommand{\winTag}[1]{{\color{rpeEdge}\bfseries\footnotesize\faCheck\ #1}\par}

\clearpage
\begin{figure*}[!t]
\section{Prompt evolution example}
\label{sec:appendix_prompt_example}
\renewcommand{\footnotesize}{\fontsize{8pt}{9.5pt}\selectfont}
\centering

\begin{tcolorbox}[
  enhanced, colback=gray!8, colframe=gray!50, boxrule=0.4pt,
  arc=2pt, left=8pt, right=8pt, top=4pt, bottom=4pt,
  before skip=0pt, after skip=4pt
]
\centering\small
\textbf{Accuracy-PE vs.\ Ranking-PE optimized prompts}:
Qwen3-VL-8B SFT $\times$ Cardiomegaly
\end{tcolorbox}

\begin{tabular}{@{}p{0.485\linewidth}@{\hspace{4pt}}p{0.485\linewidth}@{}}

\begin{aceCard}
\flawTag{Rigid hard threshold}
\promptquote{Answer ``Yes'' if the cardiac silhouette is enlarged
  (CTR $>$ 0.5 in AP/PA views) \dots\ Answer ``No'' only if the
  silhouette is clearly within normal limits \dots\ a CTR $<$ 0.5.}
Collapses borderline cases into hard Yes/No decisions, leaving the
relative ordering of ambiguous and confident cases unaddressed.
\end{aceCard}
&
\begin{rpeCard}
\winTag{Confidence calibration cues}
\promptquote{Avoid overconfidence: Even if the image appears normal,
  consider the clinical context. A patient with acute illness or
  chronic disease may have an enlarged heart that is not visually
  obvious.}
A dedicated guard against high-confidence ``No'' on visually-clean
images, encouraging graded uncertainty in ambiguous cases.
\end{rpeCard}
\\

\begin{aceCard}
\flawTag{``Yes'' bias under artifacts}
\promptquote{Do not dismiss cardiomegaly because the image is portable
  \dots\ an enlarged silhouette is still diagnostic.\\
  Do not ignore artifacts --- if the heart is partially obscured,
  assess whether the visible portion is enlarged --- if yes, it still
  counts as cardiomegaly.}
Every ``Avoid common mistakes'' clause pushes toward ``Yes''; there is no
symmetric guardrail against over-calling, so false-positives are
silently licensed.
\end{aceCard}
&
\begin{rpeCard}
\winTag{Guidance against missed findings}
\promptquote{Avoid false negatives: \dots\ Re-examine for subtle signs
  of enlargement.\\
  Avoid overconfidence: \dots\ may have an enlarged heart that is not
  visually obvious.}
The prompt directs the model to re-examine the image for subtle signs
of enlargement before ruling out cardiomegaly.
\end{rpeCard}
\\

\begin{aceCard}
\flawTag{Labs override imaging}
\promptquote{If the lab context shows signs of heart failure
  (e.g., elevated BUN/Cr, low albumin, elevated lactate, or elevated
  AST/ALT), \emph{this supports the finding even if the X-ray is
  borderline}.}
When the X-ray is borderline, lab evidence of heart failure can be the
deciding factor for a ``Yes''.
\end{aceCard}
&
\begin{rpeCard}
\winTag{Limits on using labs alone}
\promptquote{Use lab context to support or refute the finding \dots\
  lab values are \emph{not definitive} for cardiac size.\\
  Do not rely on lab values alone: Use them to support or
  contextualize the visual finding, \emph{not to replace it}.}
Labs can strengthen or weaken confidence in the visual finding. The
image remains the basis for judging cardiac enlargement.
\end{rpeCard}
\\

\begin{aceCard}
\flawTag{Stricter criteria for ``No''}
\promptquote{Answer ``Yes'' if the cardiac silhouette is enlarged
  \dots\ or if the silhouette is disproportionately large despite view or artifacts.\\
  Answer ``No'' only if the silhouette is clearly within normal limits
  \dots\ this requires clear visualization of the heart borders and a CTR $<$ 0.5.}
Apparent enlargement can trigger ``Yes'' even in the presence of view effects
or artifacts. ``No'' faces a stricter gate: clear heart borders, no visible
enlargement, and CTR below $0.5$.
\end{aceCard}
&
\begin{rpeCard}
\winTag{Ranking cues under visual ambiguity}
\promptquote{If the image is unclear or portable, \emph{do not rule
  out} cardiomegaly without strong clinical or lab evidence.\\
  Cardiomegaly is not always obvious: It may be subtle \dots\
  \emph{Do not assume absence} if the image is unclear.}
On unclear or portable images, the prompt guards against a confident
``No'' by asking the model to weigh clinical and lab evidence before
ruling out cardiomegaly.
\end{rpeCard}
\\

\begin{aceCard}
\flawTag{No structured etiology prior}
\promptquote{Note if the patient is elderly or has known heart
  disease --- these can predispose to cardiomegaly even if the X-ray
  is subtle.}
A single inline mention; no organized population-level prior to weigh
against the image when the visual evidence is ambiguous.
\end{aceCard}
&
\begin{rpeCard}
\winTag{Etiology / domain priors as a section}
\promptquote{Domain-Specific Knowledge: \dots\ Cardiomegaly can be
  associated with chronic conditions: Consider underlying causes such
  as hypertension, valvular disease, or chronic heart failure \dots}
A dedicated knowledge section gives the model an explicit cohort-level
prior to combine with imaging, a structured reasoning component absent
from the Accuracy-PE.
\end{rpeCard}

\end{tabular}

\caption{\textbf{Ranking-PE produces qualitatively different
prompts.} Both prompts are evolved from the same seed prompt under
identical infrastructure, rollout budget, and reflector; they differ
in the three Ranking-PE components (\S\ref{sec:pipeline_v2}): AUROC-based
selection, the pair-indexed scores matrix driving Pareto dominance, and
ranking-shaped feedback. The Accuracy-PE prompt builds a rigid, asymmetric decision rule around CTR.
It gives apparent enlargement several paths to ``Yes'', including support
from lab findings when the X-ray is borderline, while reserving ``No'' for
clearly visualized cases that meet its normality criteria. The Ranking-PE prompt adds safeguards against
overconfidence, guidance against missed findings and premature dismissal
on unclear images, explicit limits on using labs alone, and a structured
section on etiology. Excerpts are
from Qwen3-VL-8B SFT on Cardiomegaly and are illustrative.}
\label{fig:prompt_compare}
\end{figure*}

%
\clearpage
\begin{figure*}[p]
\centering

\begin{tcolorbox}[
  colback=gray!5, colframe=black!70, boxrule=0.4pt,
  title=\textbf{Seed prompt (shared baseline for Accuracy-PE and Ranking-PE): Qwen3-VL-8B SFT, Cardiomegaly},
  fonttitle=\small\bfseries, left=4pt, right=4pt, top=3pt, bottom=3pt,
  width=\linewidth,
]
\small
Judge if the target finding exists.
\end{tcolorbox}

\begin{minipage}[t]{\columnwidth}
\vspace{0pt}
\begin{tcolorbox}[
  colback=red!4, colframe=red!55!black, boxrule=0.4pt,
  title=\textbf{Accuracy-PE optimized prompt (Accuracy-PE, baseline): val AUROC 0.6835},
  fonttitle=\small\bfseries, left=4pt, right=4pt, top=3pt, bottom=3pt,
  width=\linewidth,
  enhanced jigsaw,
]
\scriptsize
When evaluating whether the target finding (e.g., ``Cardiomegaly'') exists in a chest X-ray, follow this structured process:

\textbf{1. Analyze the image visually:}
\begin{itemize}\itemsep0pt
\item Identify the view (AP, PA, or Portable) --- this affects how you interpret cardiac size.
\item Look for an enlarged cardiac silhouette --- specifically, assess the cardiothoracic ratio (CTR). In AP views, a CTR $>$ 0.5 suggests cardiomegaly. In PA views, a CTR $>$ 0.5 is also diagnostic, but in portable views (often AP), the heart may appear larger due to positioning --- use this as a clue, not a definitive rule.
\item Note any artifacts or equipment (e.g., ECG leads, tubes) that might obscure the cardiac borders --- if the heart is partially obscured, it may be harder to assess, but if the silhouette is clearly enlarged, it still counts.
\item Pay attention to the mediastinum --- widening may suggest cardiomegaly or other pathology, but in AP views, mediastinal widening can also be due to lung collapse or positioning.
\end{itemize}

\textbf{2. Integrate with clinical context:}
\begin{itemize}\itemsep0pt
\item Consider the patient's age, gender, and clinical profile (e.g., urgent admission, chronic conditions, lab values like BUN/Cr, electrolytes, glucose, lactate, or liver enzymes).
\item Note if the patient is elderly or has known heart disease --- these can predispose to cardiomegaly even if the X-ray is subtle.
\item If the lab context shows signs of heart failure (e.g., elevated BUN/Cr, low albumin, elevated lactate, or elevated AST/ALT), this supports the finding even if the X-ray is borderline.
\end{itemize}

\textbf{3. Avoid common mistakes:}
\begin{itemize}\itemsep0pt
\item Do not assume normal cardiac size based on a PA view --- in AP views, the heart appears larger, so cardiomegaly is more likely if the silhouette is large.
\item Do not dismiss cardiomegaly because the image is portable --- portable AP views are common in emergencies and may not be ideal for precise measurement, but an enlarged silhouette is still diagnostic.
\item Do not ignore artifacts --- if the heart is partially obscured, assess whether the visible portion is enlarged --- if yes, it still counts as cardiomegaly.
\end{itemize}

\textbf{4. Final output:}
\begin{itemize}\itemsep0pt
\item Answer ``Yes'' if the cardiac silhouette is enlarged (CTR $>$ 0.5 in AP/PA views) or if the silhouette is disproportionately large despite view or artifacts.
\item Answer ``No'' only if the silhouette is clearly within normal limits and no enlargement is visible --- this requires clear visualization of the heart borders and a CTR $<$ 0.5.
\end{itemize}

Always support your answer with a brief explanation of the visual evidence and how it aligns (or doesn't align) with the clinical context.
\end{tcolorbox}

\end{minipage}
\hfill
\begin{minipage}[t]{\columnwidth}
\vspace{0pt}
\begin{tcolorbox}[
  colback=green!4, colframe=green!45!black, boxrule=0.4pt,
  title=\textbf{Ranking-PE optimized prompt (Ranking-PE, ours): \mbox{val AUROC 0.7054}},
  fonttitle=\small\bfseries, left=4pt, right=4pt, top=3pt, bottom=3pt,
  width=\linewidth,
  enhanced jigsaw,
]
\scriptsize
You are given a chest X-ray image, a clinical profile (age, gender, race, admission type, view), and lab context (hematology, chemistry, coagulation, etc.), along with a specific finding to evaluate (e.g., ``Cardiomegaly''). Your task is to determine whether the target finding is present in the image, based on visual analysis of the chest X-ray, and to integrate this with the clinical and lab context to support your judgment.

\textbf{Visual Analysis Guidance:}
\begin{itemize}\itemsep0pt
\item Carefully examine the cardiac silhouette: Look for enlargement of the heart relative to the thoracic cavity. Cardiomegaly is typically defined as the heart occupying more than half of the thoracic cavity in AP views, or an increased transverse diameter relative to the thoracic cage.
\item Pay attention to the borders of the heart: Assess the right and left heart borders for widening, especially in the AP view where the heart appears more horizontally oriented.
\item Consider the mediastinum and lung fields: While not directly diagnostic, widened mediastinum or pleural effusions can be associated with cardiomegaly, but their absence does not rule it out.
\item Note any technical factors: Portable views may have limited detail, so subtle findings may be missed. Be cautious of over-reliance on technical limitations.
\end{itemize}

\textbf{Integration Guidance:}
\begin{itemize}\itemsep0pt
\item Use the clinical profile to contextualize the finding: Age, gender, admission type, and view (AP vs.\ PA) can influence interpretation. For example, elderly patients or those with surgical admissions may have more baseline cardiac enlargement.
\item Use lab context to support or refute the finding: Look for markers of chronic illness, renal dysfunction, electrolyte imbalances, or acute cardiac stress that may correlate with cardiomegaly. However, remember that lab values are not definitive for cardiac size.
\item Do not rely solely on the image or lab values: Combine both to form a comprehensive judgment. If the image is ambiguous or of low quality, use clinical context to guide interpretation.
\end{itemize}

\textbf{Domain-Specific Knowledge:}
\begin{itemize}\itemsep0pt
\item Cardiomegaly is not always obvious: It may be subtle, especially in portable views or in patients with normal lung fields. Do not assume absence if the image is unclear.
\item Cardiomegaly can be associated with chronic conditions: Consider underlying causes such as hypertension, valvular disease, or chronic heart failure, which may be suggested by clinical or lab data.
\item Always assess the cardiac silhouette relative to the thoracic cavity: Use the ribs and diaphragm as landmarks. The heart should not exceed half the thoracic cavity in AP views.
\end{itemize}

\textbf{Error Prevention Guidance:}
\begin{itemize}\itemsep0pt
\item Avoid false negatives: If the image is unclear or portable, do not rule out cardiomegaly without strong clinical or lab evidence. Re-examine for subtle signs of enlargement.
\item Avoid overconfidence: Even if the image appears normal, consider the clinical context. A patient with acute illness or chronic disease may have an enlarged heart that is not visually obvious.
\item Do not rely on lab values alone: Use them to support or contextualize the visual finding, not to replace it.
\end{itemize}

\textbf{Output Format:}
\begin{itemize}\itemsep0pt
\item Answer: ``Yes'' if the finding is present, ``No'' if it is not.
\item Reasoning: Provide a clear, concise explanation of your visual analysis and how it integrates with the clinical and lab context. Be specific about what you observed in the image and how it relates to the finding.
\end{itemize}
\end{tcolorbox}

\end{minipage}

\caption{\textbf{Full-text accompaniment to Figure~\ref{fig:prompt_compare}.} Word-for-word reproduction of the seed prompt (top), the Accuracy-PE-optimized prompt (left), and the Ranking-PE-optimized prompt (right) for Qwen3-VL-8B SFT $\times$ Cardiomegaly, both evolved from the same seed under identical GEPA infrastructure, rollout budget, and reflector. The runs differ in the three Ranking-PE components (\S\ref{sec:pipeline_v2}): final selection, the scores matrix that drives Pareto dominance (instance-level $\mathbf{1}[\text{correct}]$ for Accuracy-PE versus pair-indexed $r(s(x^+),s(x^-))$ with half credit for ties for Ranking-PE), and the reflector feedback. The selected excerpts in Figure~\ref{fig:prompt_compare} can be located verbatim in this figure.}
\label{fig:prompt_example_cardiomegaly}
\end{figure*}
\clearpage

\section{Ethics statement}

\noindent \textbf{Data.}
All experiments use MIMIC-IV~\citep{johnson2023mimiciv}, a retrospectively
collected, de-identified dataset accessed under PhysioNet credentialing and its
data use agreement. No new data were collected and no new human-subjects
research was conducted, so institutional review board approval is not
applicable. Because MIMIC-IV is restricted-access we do not redistribute it;
Appendix~\ref{sec:appendix_impl_data} instead documents cohort construction and
the split protocol and per-disease positive prevalence.

\noindent \textbf{Assets.}
We release no new datasets and no new model checkpoints. All backbones are
existing publicly released checkpoints used under their original licenses, and
all datasets, checkpoints and code dependencies are cited at first use.

\noindent \textbf{Intended use and risks.}
This paper contributes evaluation and optimization methodology, not a deployed
clinical system; nothing reported here is validated for clinical use. The
intended benefit is decision support optimized for the ranking behaviour
clinical workflows actually depend on, rather than for threshold accuracy, which
\S\ref{sec:intro} shows can be clinically vacuous under the class imbalance
typical of real clinical data. The corresponding risk is over-reliance on
AI-generated rankings without clinician oversight and without independent
prospective validation; we regard both as prerequisites for any deployment.
Evaluation covers three CheXpert pathologies on a single cohort, so we make no
claim about behaviour on other populations, imaging modalities, or care settings
(\S\ref{sec:limitations}).

\section{Reproducibility statement}

\S\ref{sec:setup} states the datasets, backbones, reflector configuration and
metrics used throughout. Appendix~\ref{sec:appendix_impl} supplies what is
needed to rerun both stages: the multimodal serving interface and field-level
input format (Appendix~\ref{sec:appendix_impl_mm}), the Stage~1 SFT
hyperparameters and the VE-tuned / VE-frozen distinction
(Appendix~\ref{sec:appendix_impl_sft}), and the Stage~2 search budget,
population size, reflection minibatch size and final-selection rule
(Appendix~\ref{sec:appendix_impl_pe}). Cohort construction, the
disease-selection rationale and per-disease positive prevalence are in
Appendix~\ref{sec:appendix_impl_data}. The pair-subsampling cap
\texttt{max\_pairs} and its robustness ablation are reported in
Table~\ref{tab:ablation_robustness}. Because the reflector's text feedback is
the component least likely to survive paraphrase, the exact strings it receives
are reproduced verbatim in Appendix~\ref{sec:appendix_feedback_details}
(Table~\ref{tab:error_type_rendering}) with a fully assembled example in
Figure~\ref{fig:feedback_templates}, and the seed prompt together with both
evolved prompts is reproduced word-for-word in
Appendix~\ref{sec:appendix_prompt_example}. MIMIC-IV itself cannot be
redistributed under PhysioNet credentialing; code will be released upon
publication, subject to internal review.

\clearpage
\end{document}